\documentclass{article} 
\usepackage{iclr2027_conference,times}

\usepackage{amsmath,amsfonts,bm}

\def\eqref#1{equation~\ref{#1}}

\def\1{\bm{1}}

\def\vone{{\bm{1}}}

\def\va{{\bm{a}}}
\def\vb{{\bm{b}}}

\def\ve{{\bm{e}}}

\def\vs{{\bm{s}}}

\def\vw{{\bm{w}}}
\def\vx{{\bm{x}}}
\def\vy{{\bm{y}}}

\def\mC{{\bm{C}}}

\def\mE{{\bm{E}}}

\def\mM{{\bm{M}}}

\def\mS{{\bm{S}}}

\def\mW{{\bm{W}}}
\def\mX{{\bm{X}}}

\DeclareMathAlphabet{\mathsfit}{\encodingdefault}{\sfdefault}{m}{sl}
\SetMathAlphabet{\mathsfit}{bold}{\encodingdefault}{\sfdefault}{bx}{n}

\def\sA{{\mathbb{A}}}

\def\sD{{\mathbb{D}}}

\def\sM{{\mathbb{M}}}

\def\sP{{\mathbb{P}}}

\def\sS{{\mathbb{S}}}

\newcommand{\R}{\mathbb{R}}

\usepackage{hyperref}
\usepackage{url}

\usepackage{algorithm}
\usepackage{algorithmicx}
\usepackage{algpseudocode}
\usepackage{amsmath}
\usepackage{amssymb}
\usepackage{amsfonts}
\usepackage[misc]{ifsym}    
\usepackage{tabularx}
\usepackage[table]{xcolor}
\usepackage{graphicx}
\usepackage{pdflscape}   
\usepackage{wrapfig}
\usepackage{booktabs}
\usepackage[T1]{fontenc}

\title{How to Find and Reuse Policies for Continuous Adaptation in Lifelong Reinforcement Learning}

\author{Antiquus S.~Hippocampus, Natalia Cerebro \& Amelie P. Amygdale \thanks{ Use footnote for providing further information
about author (webpage, alternative address)---\emph{not} for acknowledging
funding agencies.  Funding acknowledgements go at the end of the paper.} \\
Department of Computer Science\\
Cranberry-Lemon University\\
Pittsburgh, PA 15213, USA \\
\texttt{\{hippo,brain,jen\}@cs.cranberry-lemon.edu} \\
\And
Ji Q. Ren \& Yevgeny LeNet \\
Department of Computational Neuroscience \\
University of the Witwatersrand \\
Joburg, South Africa \\
\texttt{\{robot,net\}@wits.ac.za} \\
\AND
Coauthor \\
Affiliation \\
Address \\
\texttt{email}
}

\iclrpreprint
\author{Saptarshi Nath, Inish D'Souza \& Andrea Soltoggio \\
Department of Computer Science\\
Loughborough University\\
\texttt{\{s.nath,i.m.d-souza,a.soltoggio\}@lboro.ac.uk} \\
\And
Antonio Carta \\
Department of Computer Science \\
University of Pisa \\
\texttt{antonio.carta@unipi.it} \\
\And
Soheil Kolouri \\
Department of Computer Science \\
\texttt{soheil.kolouri@vanderbilt.edu}
}
\begin{document}

\maketitle

\begin{abstract}
    In lifelong reinforcement learning, retaining previously learned policies is not sufficient for effective transfer to a new task. Useful knowledge may be distributed across several prior policies, and its relevance may change as the learner acquires experience. One hypothesis is that task similarity can be effectively used in a continual learning setting to find and combine previously learned policies. To test it, Adaptive Mask Selection and Composition (AMSC) is designed to estimate similarity from online experience via non-parametric Wasserstein task embeddings from state-action-reward samples. The z-score-normalized sparsemax of the similarity scores are used to derive a variable-size support to periodically choose and weight policies to form a prior when learning a new task. On CT-graph and MiniGrid, AMSC achieves higher mean performance and forward transfer than the evaluated modular composition baselines while exhibiting no forgetting. Results on Continual World suggest that identifying relevant prior knowledge and determining its layer-specific composition may require additional layer-specific tuning. Ablations show that selecting relevant sources and determining how strongly to reuse them are central to these gains. Independently measured pairwise transfer is also positively associated with task-embedding similarity. These results indicate that task similarity can be an effective criterion to select and weight specific knowledge for reuse in lifelong reinforcement learning.\footnote{\url{https://github.com/Chocological45/amsc}}
\end{abstract}

\section{Introduction}
\label{sec:introduction}

Lifelong learning requires both retention and reuse of prior knowledge. Although forgetting remains a central challenge, retention alone does not ensure positive transfer to later tasks. As experience accumulates, a model must identify relevant knowledge without allowing irrelevant or conflicting knowledge to cause interference. Prior work has extensively studied retention and increasingly evaluates forward transfer \citep{rusu2016progressive, kirkpatrick2017ewc, zenke2017si, mallya2018packnet, rolnick2019clear, oswald2020continual_hypernet}, but commonly reuses shared or aggregated knowledge. Selective reuse remains comparatively underexplored.

This distinction is especially consequential in reinforcement learning (RL). Transferred knowledge affects the actions taken and the state-action distribution from which the policy learns. An unsuitable prior can bias exploration and induce negative transfer \citep{taylor2009transfer, ahn2025prevalence, Uchendu2022Jump-StartLearning}. Sparse rewards provide little early evidence that a policy is harmful \citep{gimelfarb2021contextualpolicytransfer}. Moreover, useful structure may be distributed unequally across several priors.

Existing lifelong RL methods transfer through replay, modular routing, parameter isolation, and learned composition \citep{isele2018selective, kessler2022state, powers2022neuralensembles, mendez2022modular, ben2022lifelong, zhang2024progressivecontext, malagon2024composing, hu2025continual, huang2025lifelongreinforcementlearningsimilaritydriven}. Modular approaches can preserve completed policies, but reusing the full library may admit interfering priors. Learning how strongly to reuse each policy from the target objective can identify useful policies but may converge only after their opportunity to improve early learning has passed.

Measured task similarity can instead act as a method of estimating the relevance of prior policies without optimization \citep{carroll2005transfer, taylor2009transfer, li2019contextawarepolicyreuse, gimelfarb2021contextualpolicytransfer, wu2025exploiting, nath2025policysearchretrievalcomposition}. In task-incremental lifelong RL, the availability of prior policies is determined by curriculum history, completed policies are frozen, and the number of useful sources is unknown. A suitable selector must therefore adapt to the current library. As experience is collected on a new task, both the selected knowledge and its contribution to learning may also change. We ask whether online task similarity can determine both \emph{which} policies to reuse and \emph{how strongly} they should contribute under these constraints.

\begin{wrapfigure}[19]{r}{0.5\textwidth}
    \vspace{-20pt}
    \centering
    \includegraphics[width=0.5\textwidth]{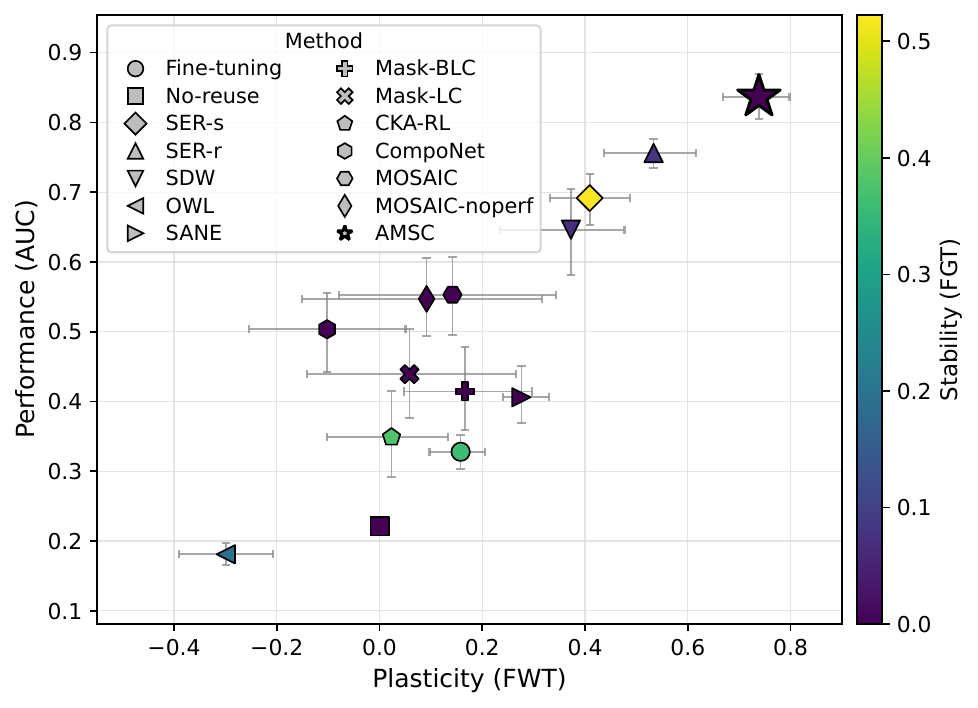}
    \caption{Transfer-retention trade-off on CT28 and MG16. Each marker reports mean FWT and AUC across seven seeds; colour denotes FGT. Error bars are 
    95\% confidence intervals.}
    \label{fig:transfer_retention_tradeoff}
\end{wrapfigure}
\textbf{Adaptive Mask Selection and Composition} (AMSC) learns task-specific policy masks over a fixed backbone network. It uses online task similarity to select a variable-size support of prior masks and determine their composition weights. Selected and current mask scores are normalized before composition with an \(\ell_2\) straight-through estimator, and only the current policy-mask scores are optimized. AMSC selects without needing task descriptions, task labels, or transfer probes.

AMSC is evaluated on CT-graph (CT28), MiniGrid (MG16), and Continual World
(CW10). It obtains AUC/FWT/FGT of \(0.94/0.94/0.00\) on CT28 and
\(0.73/0.54/0.00\) on MG16 (Figure~\ref{fig:transfer_retention_tradeoff}). In CT28 and MG16, similarity-derived weights perform comparably to learned composition weights. In CW10, similarity-derived weights underperform with respect to learned mask composition. Learning composition over a sparse selection of prior policies may provide a performance benefit. Ablations show that selecting relevant policies from the agent's library and determining how to reuse each through similarity or learned weights are critical to more effective forward transfer and successful learning in lifelong RL. Results also show that task similarity is positively associated with independently measured pairwise transfer. Lastly, results suggest that masks can be quantized with minimal forgetting and loss in AUC and FWT.


\section{Related Work}
\label{sec:related_work}

\paragraph{Selective reuse in lifelong reinforcement learning.}
Forward transfer and forgetting are distinct objectives in lifelong RL. Preserving earlier tasks does not imply faster learning on later tasks \citep{wolczyk2021continualworld, wolczyk2022transfercrl}. Prior work promotes selective reuse through sparse or compositional subnetworks \citep{sokar2022continuallearningsparsenetworks,Ostapenko2021ContinualComposition}, related and unrelated task sequences \citep{ke2020cat}, selective replay \citep{isele2018selective,huang2025lifelongreinforcementlearningsimilaritydriven}, and adaptive structure allocation \citep{lee2021sharinglessismore,dekhovich2023continual,savadikar2026cheem}. These methods select data, capacity, or structures, often to determine what knowledge is worth retaining. 

\paragraph{Modular routing and composition.}
Modular lifelong RL isolates task-specific knowledge and transfers through
routing or composition. Prior methods activate modules, select policy heads,
infer contexts, or route adapters \citep{powers2022neuralensembles,
kessler2020unclear,zhang2024progressivecontext,julian2025cable}; others compose
masks, modules, policies, or task vectors \citep{ben2022lifelong,kessler2022state,
hu2025continual,malagon2024composing}. 

\paragraph{Task similarity and optimal transport.}
Task similarity predicts transfer and supports policy selection in RL
\citep{carroll2005transfer,taylor2009transfer,li2019contextawarepolicyreuse,
gimelfarb2021contextualpolicytransfer}, while transferability methods use
representations, probe networks, likelihoods, or parameter subspaces
\citep{zamir2018taskonomydisentanglingtasktransfer,achille2019task2vec,
nguyen2020leep,lin2022trgp,wu2025exploiting}. Optimal transport compares task
distributions through OTDD and Wasserstein embeddings
\citep{courty2018learning,alvarezmelis2020otdd,liu2022wasserstein}, and supports
memory, selection, distillation, context detection, and expert diversity
\citep{wang2022distributionallyrobustmemory,wang2024inscl,xu2023manifoldexpansion,
dick2024statistical,hihn2022mixtureofvariationalexperts}. 

\section{Preliminaries}
\subsection{Problem Setting}
Consider a task-incremental lifelong RL setting with a sequence of \(K\) tasks
\(\{\mathcal{T}_k\}_{k=1}^{K}\). Each task is an MDP
\(\mathcal{T}_k=(\sS,\sA,p_k,r_k,\gamma_{\mathrm{rl}})\), where \(\sS\) and
\(\sA\) are the state and action spaces, \(p_k\) is the transition kernel,
\(r_k\) is the reward function, and \(\gamma_{\mathrm{rl}}\) is the discount
factor. Tasks arrive sequentially with known boundaries, and only the current
task environment is available for interaction. The learner trains on each task
for a fixed budget before proceeding to the next task.

A lifelong learning agent aims to maintain performance on previously learned tasks, or improve it through backward transfer, while continuing to learn new tasks through forward transfer. Forward transfer measures whether knowledge acquired from earlier tasks accelerates learning of a new task \citep{parisi2019review, khetarpal2022towards, wolczyk2021continualworld}.

\subsection{Modular Policy Masks}
\label{sec:mask_composition}
Score-based masking learns task-specific subnetworks over a shared, fixed
backbone. Supermasks in Superposition stores multiple tasks as binary masks
over the same weights \citep{wortsman2020supermasks}. Modulating Masks adapts
this mechanism to lifelong RL and introduces learned linear combinations of
the current and completed task masks \citep{ben2022lifelong}. CKA-RL is a
related modular composition method that combines and merges dense
task-specific residuals rather than binary masks \citep{hu2025continual}.

Let \(\{\mW^\ell\}_{\ell=1}^{L}\) be the fixed policy-backbone matrices and
\(\mS_k^\ell\) the real-valued score matrix for task \(k\) at masked layer
\(\ell\). The mask in the forward pass is
\begin{equation}
    \mM_k^\ell = g(\mS_k^\ell),
    \qquad
    g(x)=\1_{\{x\geq 0\}},
    \label{eq:binary_mask}
\end{equation}
where \(g\) is applied elementwise. The effective parameters are
\(\mW^\ell\odot\mM_k^\ell\). A straight-through estimator propagates
gradients through \(g\). Let \(\mS_i^{\ell,*}\) be the frozen score matrix
stored after task \(i\), and let
\(\sP(k)\subseteq\{1,\ldots,k-1\}\) contain the priors admitted for task
\(k\). A generic convex score composition is
\begin{equation}
    \mS_k^{\ell,\mathrm{comb}}
    =
    \sum_{i\in\sP(k)}
    \alpha_{k,i}^{\ell}\mS_i^{\ell,*}
    +
    \alpha_{k,k}^{\ell}\mS_k^\ell,
    \quad
    \sum_{i\in\sP(k)\cup\{k\}}\alpha_{k,i}^{\ell}=1,
    \label{eq:lc}
\end{equation}
with non-negative weights. \citet{ben2022lifelong, hu2025continual} use the complete prior library and learn layer-specific weights. For \(m>0\), \citet{ben2022lifelong} initializes the coefficients uniformly in Mask-LC or balances prior and current-mask mass in Mask-BLC. For \(m=|\sP(k)|>0\), 
\begin{equation}
    \begin{aligned}
    \text{Mask-LC:}\quad&
    \alpha_{k,k}^{\ell,(0)}
    =\alpha_{k,i}^{\ell,(0)}=\frac{1}{m+1},\\
    \text{Mask-BLC:}\quad&
    \alpha_{k,k}^{\ell,(0)}=\frac12,\qquad
    \alpha_{k,i}^{\ell,(0)}=\frac{1}{2m},
    \qquad i\in\sP(k).
    \end{aligned}
    \label{eq:initializations}
\end{equation}
When \(m=0\), the current mask has an initial weight of 1. Prior scores remain frozen, and the final composition is consolidated, \(\mS_k^{\ell,*}\leftarrow \mS_k^{\ell,\mathrm{comb}},\) for later tasks.

\section{Adaptive Mask Selection and Composition}
\label{sec:method}

AMSC consists of three operations: normalized sparsemax retrieval, scale-normalized mask
scores, and composition from the retrieval weights.

\subsection{Mask Library and Online Task Representations}

At task \(k\), AMSC initializes a new current score matrix
\(\mS_k^\ell\) and the policy library contains the frozen score matrices
\(\{\mS_i^{\ell,*}\}_{i<k,\ell=1}^{L}\) following the formulation defined in
Section~\ref{sec:mask_composition}. Let \(\sD_k\) be the current-task experience buffer and \(\mX_k^{(t)}\) a sampled batch matrix. AMSC uses an embedding map \(\psi(\mX_k^{(t)})=\ve_k^{(t)}\in\R^{d_e}\) that places online experience from different tasks in a common vector space. The main experiments use a non-parametric linear Wasserstein embedding (WTE) of state--action--reward samples \citep{liu2022wasserstein} which has been demonstrated in RL tasks \citep{nath2025policysearchretrievalcomposition}. 

Discrete actions are one-hot encoded, states are standardized within each
sampled batch, and rewards are retained as observed. Given a fixed reference
matrix \(\mX_0\) with \(R\) samples and an optimal transport matrix
\(\bm{\Gamma}^\star\), the WTE batch embedding is
\begin{equation}
    \ve_k^{(t)}
    =
    \operatorname{vec}\!\left(
        \frac{(R\bm{\Gamma}^\star)^\top\mX_k^{(t)}-\mX_0}{\sqrt{R}}
    \right).
    \label{eq:amsc_WTE}
\end{equation}
The current representation is smoothed online as
\begin{equation}
    \bar{\ve}_k^{(t)}
    =
    \operatorname{normalize}\!\left(
        \rho\bar{\ve}_k^{(t-1)}
        +(1-\rho)\ve_k^{(t)}
    \right),
    \label{eq:amsc_ema}
\end{equation}
where the first update initializes the normalized representation and
\(\rho=0.5\) thereafter. Completed task representations remain fixed.
Appendix~\ref{sec:embedding_details} formalizes the optimal transport plan and
the empirical distributions represented by \(\mX_k^{(t)}\) and \(\mX_0\).

\subsection{Sparse Retrieval}

Cosine similarity has been used to evaluate task relationships through gradient alignment, and task embeddings \citep{lopez2017gem, achille2019task2vec, yu2020gradient, nath2025policysearchretrievalcomposition}. Following this, AMSC compares the current representation of task \(k\) with every completed task using cosine similarity,
\begin{equation}
    s_{k,i}=\cos(\bar{\ve}_k,\bar{\ve}_i),
    \qquad i<k.
    \label{eq:amsc_similarity}
\end{equation}
Let \(\vs_k=[s_{k,1},\ldots,s_{k,k-1}]^\top\). AMSC retrieves
\begin{equation}
    \vw_k
    =\operatorname{sparsemax}\!\left(
        \operatorname{zscore}(\vs_k)/\tau
    \right),
    \qquad
    \sP(k)=\{i<k:w_{k,i}>0\}.
    \label{eq:amsc_retrieval}
\end{equation}
The z-score has zero mean and unit variance within the available library, and
\(\tau=1\) in all main experiments. Sparsemax projects these relative scores
onto the probability simplex and assigns exact zeros, so
\(\sum_{i\in\sP(k)}w_{k,i}=1\) \citep{martins16sparsemax}.

\subsection{Scale-Normalized Fixed Composition}

Positive rescaling of a task's scores changes its influence when scores from several tasks are combined before thresholding. Large magnitude scores become more difficult to alter. Consequently, raw score magnitude can override the linear combination weights. We therefore normalize each selected prior and current score to unit \(\ell_2\) norm in the forward pass, using a straight-through estimator to preserve the current score's gradient,
\begin{equation}
    \widehat{\mS}_i^\ell
    =
    \mS_i^\ell+
    \operatorname{sg}\!\left(
        \frac{\mS_i^\ell}
        {\max(\lVert\mS_i^\ell\rVert_F,\varepsilon_m)}
        -\mS_i^\ell
    \right),
    \label{eq:amsc_l2ste}
\end{equation}
where \(\operatorname{sg}\) denotes stop-gradient. 

Learning composition weights from the RL objective requires the agent to discover useful sources through task interaction. AMSC instead can use the same online similarity signal for both retrieval and weighting, making source contributions available at the first retrieval event without optimization. Let \(m=|\sP(k)|\). Instead of learning the composition, AMSC applies weights computed from task similarity which,
\begin{equation}
    \mC_k^\ell
    =
    \begin{cases}
        \widehat{\mS}_k^\ell,
        & m=0,\\[2mm]
        \displaystyle
        \frac{1}{m+1}\widehat{\mS}_k^\ell
        +
        \frac{m}{m+1}
        \sum_{i\in\sP(k)}
        w_{k,i}\widehat{\mS}_i^\ell,
        & m>0,
    \end{cases}
    \qquad
    \mM_k^\ell=g(\mC_k^\ell).
    \label{eq:amsc_composition}
\end{equation}
For \(m>0\), the current mask receives mass \(1/(m+1)\), while the selected priors share \(m/(m+1)\) according to their retrieval weights. These weights are shared across layers.

Each task begins without retrieved priors. Once sufficient experience is available, AMSC updates the current embedding and recomputes \eqref{eq:amsc_retrieval}--\eqref{eq:amsc_composition} between training iterations to reflect new data. At the task boundary, the final composed scores \(\mC_k^\ell\) are consolidated as \(\mS_k^{\ell,*}\leftarrow\mC_k^\ell\), frozen, and added to the reusable mask library. Appendix~\ref{sec:pseudocode} Algorithm~\ref{alg:algorithm} gives the complete training pseudo-code.

\section{Results}
\label{sec:results}

\subsection{Experimental Setup}
\label{sec:experimental_setup}

\paragraph{Benchmarks.}
We evaluate AMSC on CT28, a sparse-reward decision-tree curriculum with hierarchical tasks \citep{soltoggio2019ctgraph}; MG16, sparse-reward MiniGrid navigation tasks with shuffled observations \citep{gym_minigrid}; and CW10, Continual World continuous robotic control tasks \citep{wolczyk2021continualworld}. In CT28 and MG16, tasks within each family share a common observation space and increase in complexity. Families differ in their representations while sharing input dimensions and actions. These curricula test reuse when prior knowledge varies in relevance in relation to the next task. Details and pairwise-transfer matrices are reported in Appendix~\ref{sec:benchmarks}.

\paragraph{Baselines.}
\label{sec:baselines}
We compare against replay methods (SER \citep{isele2018selective} and SDW \citep{huang2025lifelongreinforcementlearningsimilaritydriven}), modular methods (Mask-LC/BLC \citep{ben2022lifelong}, CKA-RL \citep{hu2025continual}, CompoNet \citep{malagon2024composing}, OWL \citep{kessler2022state}, SANE \citep{powers2022neuralensembles}, MOSAIC \citep{nath2025policysearchretrievalcomposition}, and ProgNet \citep{rusu2016progressive}), fine-tuning, and learning from scratch. We also evaluate retrieval guided by independently measured pairwise transfer. Tables~\ref{tbl:results_combined} and~\ref{tbl:cw10_results} report comparisons to baselines on CT28, MG16 and CW10. Appendix~\ref{sec:hyperparameters} provides details of the hyperparameters for AMSC and baselines.

\paragraph{Protocol and Metrics.}
\label{sec:metrics_protocol}
All empirical results are evaluated with seven shared seeds. CT28 and MG16 use PPO with \(51,200\) and \(512,000\) steps per task, respectively. CW10 uses SAC with \(1\mathrm{M}\) steps per task. We report mean task area under the evaluation curve (AUC), forward transfer (FWT), and forgetting (FGT) from normalized performance between \([0,1]\), following \citep{wolczyk2021continualworld}. We report these metrics together because zero forgetting can reflect strong retention but not necessarily strong policies. Appendix~\ref{sec:hyperparameters} specifies training and evaluation hyperparameters, Appendix~\ref{sec:sensitivity_analysis} reports hyperparameter sensitivity, and Appendix~\ref{sec:metrics} reports the details of how the metrics are computed.

\subsection{Comparison to Baselines}
\label{sec:main_results}
\paragraph{Selective reuse improves learning.}
AMSC obtains strong AUC and FWT with zero measured forgetting on CT28 and MG16 (Table~\ref{tbl:results_combined}), occupying a favorable region of the transfer-retention trade-off (Figure~\ref{fig:transfer_retention_tradeoff}) with respect to baselines. AMSC achieves AUC/FWT of \(0.94/0.94\) and \(0.73/0.54\) on CT28 and MG16, respectively. From-scratch learning also avoids forgetting but obtains only \(0.22\) and \(0.23\) AUC. Surprise-prioritized SER slightly exceeds AMSC on CT28 (\(0.96/0.95\)) but incurs \(0.54\) FGT, while reservoir SER obtains \(0.63/0.20\) on MG16 with 0.07 FGT.

\begin{table}
    \caption{Evaluation on CT28 and MG16. Each cell reports the mean
    AUC/FWT/FGT with an 
    95\% confidence interval \citep{colas2018many}. AVG is computed per seed by averaging CT28 and MG16, then bootstrapping the seven paired averages. Bold and underlining denote the best and second-best non-oracle result for each metric. Significance tests are reported in Appendix~\ref{sec:significance_testing}.}
    \label{tbl:results_combined}
    \begin{center}
    \scriptsize
    \setlength{\tabcolsep}{3.5pt}
    \begin{tabularx}{\textwidth}{p{1.8cm}XXX|XXX|XXX}
        \toprule
        & \multicolumn{3}{c}{\textbf{CT28}} & \multicolumn{3}{c}{\textbf{MG16}} & \multicolumn{3}{c}{\textbf{AVG}} \\
        \cmidrule(lr){2-4} \cmidrule(lr){5-7} \cmidrule(lr){8-10}
        \textbf{Method} 
        & \textbf{AUC} \(\uparrow\) & \textbf{FWT} \(\uparrow\) & \textbf{FGT} \(\downarrow\) 
        & \textbf{AUC} \(\uparrow\) & \textbf{FWT} \(\uparrow\) & \textbf{FGT} \(\downarrow\) 
        & \textbf{AUC} \(\uparrow\) & \textbf{FWT} \(\uparrow\) & \textbf{FGT} \(\downarrow\) \\
    \midrule
        Fine-tuning
        & \(.36_{-.02}^{+.02}\) 
        & \(.22_{-.03}^{+.03}\)
        & \(.38_{-.02}^{+.02}\)
        
        & \(.30_{-.05}^{+.05}\) 
        & \(.10_{-.09}^{+.08}\) 
        & \(.34_{-.07}^{+.08}\)
        
        & \(.33_{-.02}^{+.02}\) 
        & \(.16_{-.06}^{+.05}\)
        & \(.36_{-.03}^{+.03}\) \\
        
        From-scratch
        & \(.22_{-.01}^{+.01}\) 
        & \(.00_{-.00}^{+.00}\)
        & \(\underline{.00}_{-.00}^{+.00}\)
        
        & \(.23_{-.02}^{+.03}\) 
        & \(.00_{-.00}^{+.00}\)
        & \(\mathbf{.00}_{-.00}^{+.00}\)
        
        & \(.22_{-.01}^{+.01}\) 
        & \(.00_{-.00}^{+.00}\)
        & \(\mathbf{.00}_{-.00}^{+.00}\) \\
        
        \midrule
        SER-s
        & \(\mathbf{.96}_{-.03}^{+.02}\) 
        & \(\mathbf{.95}_{-.03}^{+.03}\)
        & \(.54_{-.18}^{+.17}\)
        
        & \(.42_{-.07}^{+.06}\)
        & -\(.13_{-.17}^{+.17}\)
        & \(.51_{-.09}^{+.10}\)
        
        & \(.69_{-.04}^{+.03}\)
        & \(.41_{-.08}^{+.08}\)
        & \(.52_{-.11}^{+.09}\) \\
        
        SER-r
        & \(.88_{-.05}^{+.05}\)
        & \(.87_{-.06}^{+.06}\)
        & \(.09_{-.07}^{+.08}\)
        
        & \(\underline{.63}_{-.03}^{+.04}\) 
        & \(.20_{-.20}^{+.14}\)
        & \(\underline{.07}_{-.09}^{+.11}\)
        
        & \(\underline{.76}_{-.02}^{+.02}\) 
        & \(\underline{.53}_{-.10}^{+.08}\)
        & \(.08_{-.06}^{+.09}\) \\
        
        SDW
        & \(.76_{-.10}^{+.11}\) 
        & \(.73_{-.13}^{+.13}\)
        & \textbf{-}\(\mathbf{.01}_{-.01}^{+.01}\)
        
        & \(.53_{-.06}^{+.05}\) 
        & \(.02_{-.18}^{+.12}\)
        & \(.15_{-.12}^{+.13}\)
        
        & \(.65_{-.07}^{+.06}\)
        & \(.37_{-.14}^{+.10}\)
        & \(\underline{.07}_{-.06}^{+.06}\) \\
        
        \midrule
        OWL
        & \(.15_{-.01}^{+.01}\) 
        & -\(.31_{-.20}^{+.17}\)
        & \(.18_{-.04}^{+.03}\)
        
        & \(.21_{-.02}^{+.03}\)
        & -\(.29_{-.19}^{+.18}\)
        & \(.22_{-.02}^{+.03}\)
        
        & \(.18_{-.02}^{+.02}\)
        & -\(.30_{-.09}^{+.09}\)
        & \(.20_{-.02}^{+.02}\) \\
        
        SANE
        & \(.46_{-.03}^{+.02}\) 
        & \(.37_{-.03}^{+.03}\)
        & \(\underline{.00}_{-.00}^{+.00}\)
        
        & \(.36_{-.08}^{+.09}\)
        & \(.18_{-.09}^{+.11}\)
        & \(\mathbf{.00}_{-.00}^{+.00}\)
        
        & \(.41_{-.04}^{+.04}\)
        & \(.28_{-.04}^{+.05}\)
        & \(\mathbf{.00}_{-.00}^{+.00}\) \\
        
        Mask-BLC
        & \(.27_{-.02}^{+.02}\) 
        & \(.10_{-.06}^{+.06}\)
        & \(\underline{.00}_{-.00}^{+.00}\)
        
        & \(.55_{-.10}^{+.10}\) 
        & \(\underline{.23}_{-.19}^{+.21}\)
        & \(\mathbf{.00}_{-.00}^{+.00}\)
        
        & \(.41_{-.06}^{+.06}\)
        & \(.17_{-.12}^{+.13}\)
        & \(\mathbf{.00}_{-.00}^{+.00}\) \\
        
        Mask-LC
        & \(.48_{-.03}^{+.03}\) 
        & \(.35_{-.07}^{+.06}\)
        & \(\underline{.00}_{-.00}^{+.00}\)
        
        & \(.40_{-.13}^{+.14}\) 
        & -\(.24_{-.37}^{+.41}\)
        & \(\mathbf{.00}_{-.00}^{+.00}\)
        
        & \(.44_{-.06}^{+.06}\)
        & \(.06_{-.20}^{+.21}\)
        & \(\mathbf{.00}_{-.00}^{+.00}\) \\
        
        CKA-RL
        & \(.36_{-.04}^{+.05}\)
        & \(.25_{-.05}^{+.05}\)
        & \(.36_{-.05}^{+.05}\)
        
        & \(.34_{-.09}^{+.11}\)
        & -\(.21_{-.24}^{+.20}\)
        & \(.40_{-.12}^{+.14}\)
        
        & \(.35_{-.06}^{+.07}\) 
        & \(.02_{-.13}^{+.11}\)
        & \(.38_{-.06}^{+.08}\) \\
        
        CompoNet
        & \(.69_{-.10}^{+.08}\) 
        & \(.64_{-.12}^{+.09}\)
        & \(\underline{.00}_{-.00}^{+.00}\)
        
        & \(.31_{-.08}^{+.07}\) 
        & -\(.85_{-.25}^{+.24}\)
        & \(\mathbf{.00}_{-.00}^{+.00}\)
        
        & \(.50_{-.06}^{+.05}\) 
        & -\(.10_{-.15}^{+.15}\)
        & \(\mathbf{.00}_{-.00}^{+.00}\) \\
        
        MOSAIC
        & \(.81_{-.03}^{+.03}\) 
        & \(.79_{-.03}^{+.03}\)
        & \(\underline{.00}_{-.00}^{+.00}\)
        
        & \(.29_{-.11}^{+.11}\) 
        & -\(.50_{-.45}^{+.40}\)
        & \(\mathbf{.00}_{-.00}^{+.00}\)
        
        & \(.55_{-.06}^{+.05}\) 
        & \(.14_{-.22}^{+.20}\)
        & \(\mathbf{.00}_{-.00}^{+.00}\) \\
        
        MOSAIC $\neg\mathbb{I}_{\mathrm{perf}}$
        & \(.81_{-.03}^{+.03}\) 
        & \(.78_{-.03}^{+.03}\)
        & \(\underline{.00}_{-.00}^{+.00}\)
        
        & \(.29_{-.11}^{+.13}\) 
        & -\(.60_{-.50}^{+.47}\)
        & \(\mathbf{.00}_{-.00}^{+.00}\)
        
        & \(.55_{-.05}^{+.06}\)
        & \(.09_{-.24}^{+.22}\)
        & \(\mathbf{.00}_{-.00}^{+.00}\) \\
        
        \midrule
        Transfer oracle
        & \( .82_{-.01}^{+.00} \)
        & \( .72_{-.13}^{+.07} \)
        & \( .00_{-.00}^{+.00} \)
        
        & \( .66_{-.09}^{+.08} \)
        & \( .54_{-.10}^{+.10} \)
        & \( .00_{-.00}^{+.00} \)
        
        & \( .74_{-.04}^{+.04} \)
        & \( .63_{-.09}^{+.08} \)
        & \(\mathbf{.00}_{-.00}^{+.00} \) \\
        
        \textbf{AMSC}
        & \(\underline{.94}_{-.03}^{+.02}\) 
        & \(\underline{.94}_{-.03}^{+.02}\)
        & \(\underline{.00}_{-.00}^{+.00}\)
        
        & \(\mathbf{.73}_{-.06}^{+.06}\)
        & \(\mathbf{.54}_{-.14}^{+.12}\)
        & \(\mathbf{.00}_{-.00}^{+.00}\)
        
        & \(\mathbf{.84}_{-.03}^{+.03}\)
        & \(\mathbf{.74}_{-.07}^{+.06}\)
        & \(\mathbf{.00}_{-.00}^{+.00}\) \\
        \bottomrule
    \end{tabularx}
    \end{center}
\end{table}
\paragraph{Comparison with modular reuse.}
AMSC achieves the highest mean AUC and FWT among modular methods on CT28 and MG16 (Table~\ref{tbl:results_combined}). These methods differ in how they reuse prior knowledge: Mask-LC/BLC learn weights over all prior masks, CompoNet learns to combine policy-unit outputs, and MOSAIC selects masks through absolute similarity and performance gates before optimizing composition. AMSC instead periodically updates both its selected sources and their weights using library-relative similarity. On CT28, AMSC achieves AUC/FWT of \(0.94/0.94\), compared with \(0.81/0.79\) for MOSAIC and \(0.69/0.64\) for CompoNet. On MG16, both baselines have negative FWT, whereas AMSC achieves \(0.54\). AMSC's gains over Mask-LC, CompoNet, and MOSAIC are significant on both metrics and curricula (\(p\leq0.023\)). Its MG16 advantage over Mask-BLC (\(0.73/0.54\) versus \(0.55/0.23\), \(p=0.016\) and \(p=0.016\), respectively). These results associate explicit, relative selection and composition with stronger reuse than dense composition or fixed absolute selection.

\begin{wraptable}[11]{r}{0.36\textwidth}
    \vspace{-40pt}
    \caption{CW10 performance across seven seeds. Entries report means with 
    95\% confidence intervals.} 
    \label{tbl:cw10_results}
    \begin{center}
    \scriptsize
    \setlength{\tabcolsep}{2.2pt}
    \begin{tabular}{lccc}
        \toprule
        \textbf{Method}
        & \textbf{AUC}
        & \textbf{FWT}
        & \textbf{FGT} \\
        \midrule

        Fine-tuning
        & \(.34_{-.18}^{+.17}\)
        & -\(7.12_{-13.38}^{+7.20}\)
        & \(.39_{-.21}^{+.19}\) \\

        From-scratch
        & \(.41_{-.04}^{+.04}\)
        & \(.00_{-.00}^{+.00}\)
        & \(.00_{-.00}^{+.00}\) \\

        PNN
        & \(.34_{-.03}^{+.02}\)
        & -\(.31_{-.14}^{+.14}\)
        & \(.00_{-.00}^{+.00}\) \\

        Mask-LC
        & \(.47_{-.06}^{+.06}\)
        & -\(.22_{-.40}^{+.34}\)
        & \(.00_{-.00}^{+.00}\) \\

        Mask-BLC
        & \(.53_{-.04}^{+.04}\)
        & \(.12_{-.20}^{+.19}\)
        & \(.00_{-.00}^{+.00}\) \\

        CKA-RL
        & \(.39_{-.14}^{+.10}\)
        & -\(2.12_{-2.24}^{+1.76}\)
        & \(.00_{-.00}^{+.00}\) \\

        MOSAIC
        & \(.26_{-.03}^{+.04}\)
        & -\(4.76_{-6.41}^{+3.74}\)
        & \(.00_{-.00}^{+.00}\) \\
        \midrule
        AMSC
        & \(.48_{-.05}^{+.05}\)
        & -\(1.21_{-.98}^{+.79}\)
        & \(.00_{-.00}^{+.00}\) \\

        \bottomrule
    \end{tabular}
    \end{center}
\end{wraptable}

\paragraph{Early learning across reuse mechanisms.}
Over the first 20\% of CT28 tasks 1--27, AMSC obtains \(\mathrm{AUC}_{20}=0.86\), compared with \(0.35\) for MOSAIC (\(p=0.016\)). The first positive training reward appears after \(2.34\mathrm{k}\) versus \(5.83\mathrm{k}\) steps (\(p=0.016\)). SER-surprise reaches \(0.91\) early AUC and SER-reservoir \(0.86\); neither differs significantly from AMSC on this metric. Thus AMSC's early advantage is clear relative to MOSAIC, while replay can match or exceed its early learning. Appendix~Tables~\ref{tbl:ct28_early_acquisition_summary} and~\ref{tbl:ct28_early_acquisition} report aggregate and per-task results.

\subsection{Limits of Similarity Weighting}
\label{sec:cw10_analysis}

\paragraph{Does fixed similarity weighting extend to continuous control?}
On CW10, AMSC obtains \(0.48\) AUC with zero measured forgetting, but its FWT is \(-1.21\); Mask-BLC obtains \(0.53\) AUC and \(0.12\) FWT (Table~\ref{tbl:cw10_results}). Their composition weights overlap on some tasks, yet AMSC often allocates more mass to prior masks later in the curriculum. At task 9, mean actor prior mass is \(0.78\) for
AMSC versus \(0.46\) for Mask-BLC (Appendix~\ref{sec:cw10_analysis_extended}). AMSC applies the same similarity-derived proportions across masked layers, whereas Mask-BLC learns layer-specific weights. This difference suggests that identifying relevant prior policies and determining how to compose their knowledge are distinct challenges, even when learned composition addresses both through a single set of weights. 

\subsection{Task Representation Sensitivity}
\label{sec:representation_sensitivity}
One question is whether the performance of AMSC is dependent on the specific representation approach, or whether alternatives to WTE can be used with similar performance. Table~\ref{tbl:embedding_study_fixed_sparsemax} compares WTE \citep{liu2022wasserstein}, sliced-Wasserstein embeddings \citep{naderializadeh2021pswe}, an empirical-Fisher adaptation of Task2Vec \citep{achille2019task2vec}, recurrent encoders motivated by PEARL and VariBAD \citep{pmlr-v97-rakelly19a,Zintgraf2019VariBAD:Meta-Learning}, and FID-inspired Fr\'echet moments \citep{heusel2017gans}. All comparisons receive complete \((s,a,r,s',d)\) as Task2Vec-style and learned context encoder variants require the reward and next state as part of the objective. The main AMSC experiments use only \((s,a,r)\) (also reported in Table~\ref{tbl:embedding_study_fixed_sparsemax} for reference).

\begin{wraptable}[13]{r}{0.45\textwidth}
    \vspace{-30pt}
    \caption{Task-representation sensitivity of AMSC. Reports AUC/FWT with 95\% confidence intervals across seven seeds. T denotes complete transitions; R and L denote random and online-learned. Bold and underlining indicate the best and second-best results.}
    \label{tbl:embedding_study_fixed_sparsemax}
    \begin{center}
    \scriptsize
    \setlength{\tabcolsep}{3.5pt}
    \begin{tabular}{lcc}
    \toprule
    \textbf{Method}
    & \textbf{CT28 AUC\(\uparrow\)/FWT\(\uparrow\)}
    & \textbf{MG16 AUC\(\uparrow\)/FWT\(\uparrow\)} \\
    \midrule
    WTE (T)
    & \(\underline{.93}_{-.04}^{+.04}/ \underline{.92}_{-.04}^{+.04}\)
    & \(.61_{-.08}^{+.07}/.38_{-.14}^{+.12}\) \\
    SWE (T)
    & \(.92_{-.03}^{+.04}/.91_{-.05}^{+.05}\)
    & \(.67_{-.04}^{+.04}/.39_{-.16}^{+.15}\) \\
    Fisher (T2V)
    & \(.87_{-.07}^{+.06}/.85_{-.10}^{+.08}\)
    & \(.45_{-.10}^{+.12}/.21_{-.18}^{+.17}\) \\
    Context (R)
    & \(.92_{-.03}^{+.03}/.91_{-.04}^{+.03}\)
    & \(.64_{-.07}^{+.06}/.41_{-.12}^{+.10}\) \\
    Context (L)
    & \(.90_{-.03}^{+.03}/.87_{-.06}^{+.06}\)
    & \(.44_{-.10}^{+.09}/.14_{-.28}^{+.21}\) \\
    Fr\'echet
    & \(.90_{-.03}^{+.03}/.87_{-.07}^{+.06}\)
    & \(\underline{.70}_{-.07}^{+.07}/
       \underline{.47}_{-.16}^{+.13}\) \\
    \midrule
    \textbf{WTE (SAR)}
    & \(\mathbf{.94_{-.03}^{+.02}}/
       \mathbf{.94_{-.03}^{+.02}}\)
    & \(\mathbf{.73_{-.06}^{+.06}}/
       \mathbf{.54_{-.14}^{+.12}}\) \\
    \bottomrule
    \end{tabular}
    \end{center}
\end{wraptable}
\paragraph{Representation choice is task-dependent but AMSC remains agnostic.}
Among the variants, WTE--transition has the highest mean AUC/FWT on CT28 (\(0.93/0.92\)), while Fr\'echet moments lead on MG16 (\(0.70/0.47\)). SWE and random recurrent features also yield positive transfer on both curricula. AMSC therefore accommodates several representations, although performance depends on the specific distributions being measured and specific configurations used and no single approach is universally best.  

\subsection{Retrieval, Composition, and Causal Transfer}
\label{sec:selection_analysis}
We test whether similarity identifies useful prior knowledge using independently measured pairwise transfer. For each source--target pair, we measure \(\Delta\mathrm{AUC}\) (target task AUC with a source minus AUC without it). 
\begin{figure}
    \centering
    \includegraphics[width=1.0\textwidth]{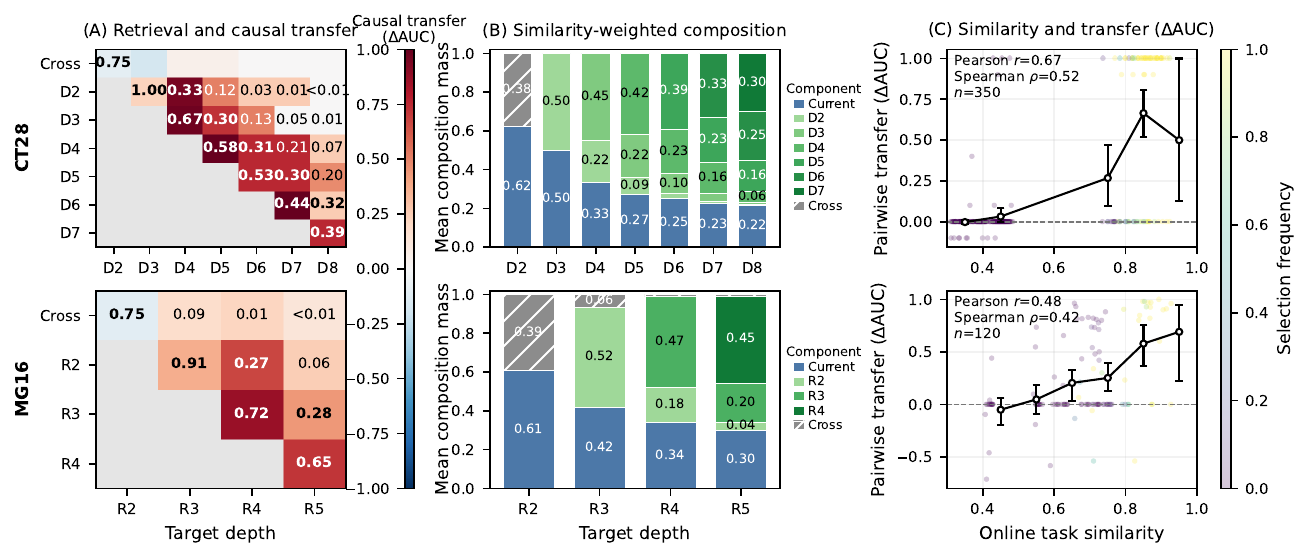}
    \caption{AMSC on CT28 (top) and MG16 (bottom). Same-family source policies share the target task's observation space. (A) Mean pairwise \(\Delta\mathrm{AUC}\) by source category. Annotations give the mean total sparsemax retrieval mass assigned to that category. (B) Mean composition mass after applying Equation~\ref{eq:amsc_composition} to 
    selected sources. (C) Online similarity versus pairwise transfer; color denotes selection frequency and the line shows bin means with target-cluster 
    95\% confidence intervals. CT28 includes depths D4--D8, whereas MG16 includes all causal non-self pairs. Retrieval uses seven seeds; pairwise transfer uses one source-policy seed.}
    \label{fig:selection_composition}
\end{figure}

\paragraph{Does task similarity rank useful priors?}
Figure~\ref{fig:selection_composition}C presents the correlation between pairwise transfer and similarity. Online similarity correlates with pairwise transfer on CT28 depths 4--8 (Pearson \(r=0.67\), 95\% CI \([0.57, 0.77]\) and MG16 (\(r=0.48, [0.31, 0.62]\)) with Spearman \(\rho=0.52\) and \(0.42\). These correlations suggest that similarity is informative of which prior masks are likely to transfer.


\paragraph{Does retrieval exploit this ranking?}
We define a pair as regularly selected if it is retrieved in at least half of its available selection events. On CT28, regularly selected pairs have mean \(\Delta\mathrm{AUC}=0.66\), versus \(0.02\) for other pairs. These pairs cover 81\% of positive-transfer pairs and include no negative-transfer pairs. On MG16, corresponding means are \(0.49\) and \(0.15\), but retrieval covers only \(19\) of \(47\) positive-transfer pairs as a result of the sparse selection.

\paragraph{How is reuse allocated within priors?}
On CT28, support grows from one prior at depth 3 to approximately four at depth 8, reducing current-mask mass from \(0.50\) to \(0.22\). At depth 8, predecessors at depths 7 and 6 receive \(0.30\) and \(0.25\) of the mass. MG16 shows a similar pattern: at depth-5 (R5), the current mask and depth-4 (R4) and depth-3 (R3) masks receive \(0.30\), \(0.45\), and \(0.20\) and below \(0.01\) to selected cross-family masks. Reuse therefore concentrates on related hierarchical sub-task masks obtained earlier in learning, while retaining a current-task component (Figure~\ref{fig:selection_composition}B).



\begin{wraptable}[14]{r}{0.47\textwidth}
    \vspace{-43pt}
    \caption{Ablation of AMSC's selection and composition mechanism. All runs
    use WTE embeddings of state--action--reward tuples. Values are means with
    95\% confidence intervals. All configurations obtain zero FGT.}
    \label{tbl:amsc_composition_ablation}
    \begin{center}
    \scriptsize
    \setlength{\tabcolsep}{3.5pt}
    \begin{tabular}{lcc}
        \toprule
        \textbf{Configuration}
        & \textbf{CT28 AUC$\uparrow$/FWT$\uparrow$} 
        & \textbf{MG16 AUC$\uparrow$/FWT$\uparrow$} \\
        \midrule
        No Sim.-\(\alpha\)
        & \(.90_{-.03}^{+.03}/.88_{-.04}^{+.04}\)
        & \(.61_{-.10}^{+.10}/.41_{-.15}^{+.15}\) \\
        No L2-STE
        & \(.89_{-.03}^{+.03}/.86_{-.03}^{+.03}\)
        & \(.59_{-.10}^{+.09}/.21_{-.29}^{+.26}\) \\
        No Periodicity
        & \(.87_{-.05}^{+.04}/.85_{-.06}^{+.05}\)
        & \(.63_{-.13}^{+.11}/.42_{-.15}^{+.14}\) \\
        \midrule
        Top-1
        & \(.88_{-.04}^{+.04}/.86_{-.04}^{+.05}\)
        & \(.63_{-.08}^{+.09}/.39_{-.16}^{+.16}\) \\
        Learned-$\alpha$
        & \(\underline{.92}_{-.03}^{+.03}/\underline{.91}_{-.04}^{+.04}\)
        & \(.60_{-.11}^{+.11}/.33_{-.14}^{+.15}\) \\
        Random
        & \(.35_{-.02}^{+.02}/.13_{-.06}^{+.07}\)
        & \(.36_{-.06}^{+.05}/.08_{-.13}^{+.13}\) \\
        Transf. Oracle
        & \(.82_{-.01}^{+.00}/.72_{-.13}^{+.07}\)
        & \(\underline{.66}_{-.09}^{+.08}/\underline{.54}_{-.10}^{+.10}\) \\
        \midrule
        \textbf{AMSC}
        & \(\mathbf{.94}_{-.03}^{+.02}/\mathbf{.94}_{-.03}^{+.02}\)
        & \(\mathbf{.73_{-.06}^{+.06}}/\mathbf{.54_{-.14}^{+.12}}\) \\
        \bottomrule
    \end{tabular}
    \end{center}
\end{wraptable}
\subsection{Ablation Studies}
\label{sec:amsc_ablations}
\paragraph{Controls.}
Table~\ref{tbl:amsc_composition_ablation} tests whether the performance gains of AMSC are attributed to the combination of components. \emph{Random} permutes selected priors while preserving how many are selected and their specific similarity-derived weights. \emph{Learned}-\(\alpha\) optimizes the retrieved weights, whereas \emph{No Sim.}-\(\alpha\) tests a uniform composition over selected prior masks. \emph{No Periodicity} retains the first retrieval for the entire learning duration. \emph{Top-1} retains the highest-similarity prior after sparsemax selection. \emph{Transf. Oracle} uses independent pairwise \(\Delta\mathrm{AUC}\) measurements to guide retrieval of multiple masks but is not an upper bound on joint composition.

\paragraph{Does source relevance matter beyond sparsity?}
Randomizing which prior masks are selected significantly reduces AUC/FWT to \(0.35/0.13\) on CT28 and \(0.36/0.08\) on MG16 (\(p=0.016\)). Because support size and weights are preserved, these reductions show that sparse composition alone is insufficient and selecting relevant priors is central to the gains.

\paragraph{Do retrieval updates and multiple sources improve transfer?}
Freezing the first retrieval lowers AUC/FWT to \(0.87/0.85\) on CT28 and \(0.63/0.42\) on MG16. These decreases suggest that revising the support and weights as experience accumulates contributes beyond the initial selection. In AMSC, selection remains stable throughout learning. Consecutive-selection Jaccard similarity is \(0.94\)--\(1.00\) on CT28 and \(0.97\)--\(1.00\) on MG16. AMSC successfully identifies relevant prior masks early on in learning then gradually refines its selection. Restricting retrieval to one prior mask results in \(0.88/0.86\) and \(0.63/0.39\), respectively. Since total prior mass is unchanged, this comparison supports composing multiple sources rather than assigning the entire reuse allocation to the highest-ranked prior.

\paragraph{Transfer-informed retrieval.}
Transfer oracle also obtains positive FWT, obtaining AUC/FWT of \(0.82/0.72\) on CT28 and \(0.66/0.54\) on MG16. It nevertheless falls below AMSC on CT28 and matches only its FWT on MG16. Selection is guided by pair-wise transfer, therefore it does not provide an upper bound on joint composition.

\paragraph{Do composition weights need to be learned?}
Learning weights within the retrieved support yields AUC/FWT of \(0.92/0.91\) on CT28 and \(0.60/0.33\) on MG16, compared with \(0.94/0.94\) and \(0.73/0.54\) for AMSC. Optimization thus provides no improvement in mean performance in these experiments. Replacing similarity-derived source proportions with uniform weights also lowers performance, to \(0.90/0.88\) and \(0.61/0.41\). These comparisons suggest that similarity provides useful information about how strongly to reuse selected masks, beyond determining which masks enter the composition; however, task similarity in AMSC cannot determine how knowledge should be uniquely composed at every layer of the network as indicated by the earlier CW10 results.

\paragraph{Does mask-score normalization matter?}
Composition weights determine source contributions only relative to the magnitudes of the scores being combined. Removing L2-STE lowers AUC/FWT to \(0.89/0.86\) on CT28 and \(0.59/0.21\) on MG16. These results support normalizing scores before composition so that differences in score magnitude do not override the intended similarity-derived allocation.

\subsection{Addressing Per-Task Mask Memory Costs}
\label{sec:mask_quantization}
Continuous-value masks are memory intensive. We test whether masks can be quantized with minimal performance loss. Ternary quantization maps scores to \(\{-1, 0, 1\}\), preserving their sign while suppressing near-zero scores. 
On CT28, ternary quantization achieves AUC/FWT/FGT of \(0.94/0.91/0.04\) compared to \(0.941/0.935/0.000\) for Float32 masks. Ternary scores also reduce per-task storage by \(16\times\) relative to Float32, while retaining better AUC/FWT (Appendix~\ref{sec:mask_score_analysis}).

\section{Discussion and Future Work}
\label{sec:discussion}
The results suggest that task similarity can guide effective policy reuse without directly optimizing retrieval or composition weights when relevant prior knowledge is available. On CT28 and MG16, AMSC achieves competitive AUC and FWT relative to replay methods while avoiding forgetting through per-task masks. The random-selection ablation indicates that sparsity alone is insufficient. Periodic retrieval further improves reuse as more experience is collected and task representations are updated.

The differing performance of transition embeddings and Fr\'echet moments also suggests that representation choice depends on the distributions. Future work could evaluate combinations of priors beyond pairwise transfer to determine how reliably embedding similarity predicts joint transfer. The CW10 results highlight a distinction between identifying relevant knowledge and determining how to reuse it. Similarity-derived weights shared across layers may be too restrictive for effective composition. Future work could test whether task embeddings can guide layer-specific reuse or whether similarity-based selection could be combined with learned composition, including an adaptive balance between prior and current knowledge. Another direction is determining when an agent should abstain from reuse when no relevant prior knowledge exists.

This work examines how sparsity and selectivity can enable effective forward transfer as the policy library grows, while managing that growth remains a complementary challenge. Since available knowledge influences what can be effectively learned, consolidation and compression should be evaluated for their effects on future transfer as well as retained performance. Finally, extending AMSC beyond known task boundaries requires detecting distribution changes and deciding when to retrieve, adapt, or create a new mask.

\section{Conclusion}
\label{sec:conclusion}
This work introduced Adaptive Mask Selection and Composition (AMSC) to address how lifelong agents can selectively reuse prior knowledge when learning new tasks. AMSC uses task similarity to periodically select and compose a sparse set of policy masks, without directly optimizing selection or composition. On CT28 and MG16, it outperforms modular baselines in mean AUC and forward transfer and remains competitive with replay methods while retaining prior policies without forgetting. These findings show that task similarity can provide an effective alternative to learning compositions, with ablations supporting the importance of relevant selection and updating reuse during learning. The advantage is not universal, however, as the CW10 results suggest that identifying relevant prior knowledge and adapting its composition across layers may require different mechanisms. The effectiveness of similarity-guided reuse depends on how prior knowledge is composed per-layer. Together, the results position task similarity as a practical approach for selective reuse.

\subsubsection*{Acknowledgments}
This material is based upon work supported by the Defense Advanced Research Projects Agency (DARPA) under contract No. HR001121901 (Shared Experience Lifelong Learning).

\bibliography{iclr2027_conference}
\bibliographystyle{iclr2027_conference}

\appendix
\section{Mask-Score Sensitivity and Quantized Storage}
\label{sec:mask_score_analysis}

\paragraph{Motivation and relation to binary supermasks.}
AMSC represents task \(i\) in layer \(\ell\) by a real-valued score tensor
\(S_i^{\ell}\) over a frozen weight tensor \(W^{\ell}\). The forward
pass uses a binary gate
\begin{equation}
    M_i^{\ell}=\mathbb{I}[S_i^{\ell}\ge0],
    \qquad
    h^{\ell+1}=\sigma\!\left(
        (W^{\ell}\odot M_i^{\ell})h^{\ell}
    \right),
    \label{eq:binary_mask_gate}
\end{equation}
and propagates gradients to \(S_i^{\ell}\) with a straight-through estimator
(STE). This follows \citet{wortsman2020supermasks, ben2022lifelong}, which optimize a subnetwork over a frozen backbone. Related signed-supermask work allows connections to be kept, hidden, or inverted \citep{koster2022signing_sm}.
AMSC instead keeps the deployed gate in Equation~\ref{eq:binary_mask_gate} binary. It retains continuous scores for the linear combination before the gate is applied in the forward pass on the resulting mask. Score magnitude can therefore affect composition even though only the score sign determines an individual task's subnetwork.

\citet{wortsman2020supermasks} also superimposes learned binary masks,
but optimizes their mixture to infer a task at evaluation time. AMSC composes completed policies during target-task learning, then consolidates the composition as the next reusable policy. The following experiments test whether continuous score information is required for this transfer mechanism and whether completed masks can instead be stored in quantized form.

\paragraph{Magnitude-controlled perturbations.}
We first test whether score magnitude identifies stable gates. For a score
tensor with \(D_{\ell}\) entries, define its root-mean-square (RMS) scale as
\begin{equation}
    r_i^{\ell}
    =\sqrt{\frac{1}{D_{\ell}}
        \sum_{j=1}^{D_{\ell}}(S_{i,j}^{\ell})^2}.
    \label{eq:mask_score_rms}
\end{equation}
For a selected index set \(\mathcal I\), each score receives independent
zero-mean Gaussian noise,
\begin{equation}
    \widetilde S_{i,j}^{\ell}
    =S_{i,j}^{\ell}+\epsilon_{i,j}^{\ell},
    \qquad
    \epsilon_{i,j}^{\ell}\sim
    \mathcal N\!\left(0,(\alpha r_i^{\ell})^2\right),
    \quad j\in\mathcal I.
    \label{eq:mask_score_perturbation}
\end{equation}
Zero-mean noise preserves each score in expectation. RMS scaling makes
\(\alpha\) dimensionless and comparable across layers with different score
scales. Thus, \(\alpha=0.5\) sets the noise standard deviation to one half of
the corresponding layer RMS. For a nonzero score \(s\), the probability that
Equation~\ref{eq:binary_mask_gate} changes state is
\begin{equation}
    \Pr\!\left[
        \mathbb I[s+\epsilon>0]\ne\mathbb I[s>0]
        \mid s
    \right]
    =\Phi\!\left(-\frac{|s|}{\alpha r_i^{\ell}}\right),
    \label{eq:mask_gate_flip_probability}
\end{equation}
where \(\Phi\) is the standard-normal cumulative distribution function.
Equation~\ref{eq:mask_gate_flip_probability} predicts frequent gate changes
near zero and rapidly fewer changes as normalized magnitude increases.

The structural sweep perturbs the lowest-magnitude, random, or
highest-magnitude fraction of each tensor. We evaluate fractions
\(f\in\{0.10,0.25,0.50\}\) and scales
\(\alpha\in\{0.01,0.05,0.10,0.25,0.50,1.00\}\) using final checkpoints from
the seven CT28 and MG16 seeds. Matched-state policy analysis perturbs
actor-relevant trunk and head scores and measures deterministic-action
disagreement. Fresh-rollout return estimates use three episodes per task and
serve as a diagnostic rather than an equivalence test.

\begin{figure*}[t]
    \centering
    \begin{minipage}{0.49\textwidth}
        \centering
        \includegraphics[width=\linewidth]{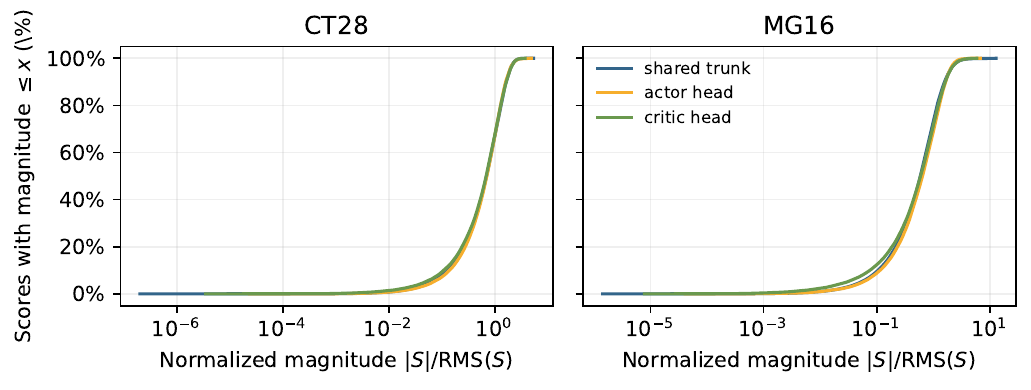}
        \textbf{(a) Score-magnitude distributions.}
    \end{minipage}
    \hfill
    \begin{minipage}{0.49\textwidth}
        \centering
        \includegraphics[width=\linewidth]{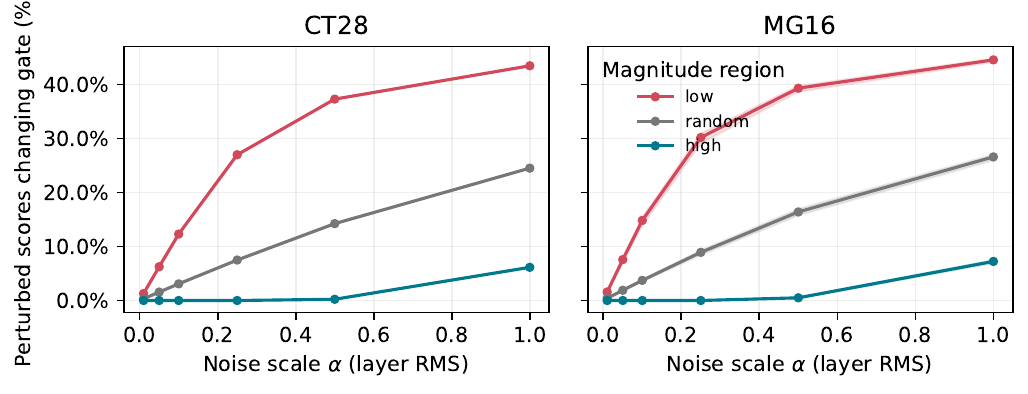}
        \textbf{(b) Gate changes under perturbation.}
    \end{minipage}

    \vspace{0.7em}
    \begin{minipage}{0.49\textwidth}
        \centering
        \includegraphics[width=\linewidth]{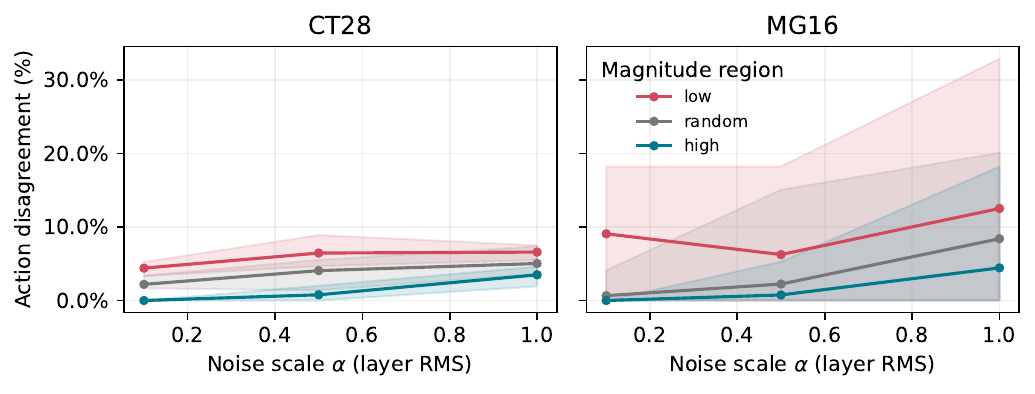}
        \textbf{(c) Action disagreement on matched states.}
    \end{minipage}
    \hfill
    \begin{minipage}{0.49\textwidth}
        \centering
        \includegraphics[width=\linewidth]{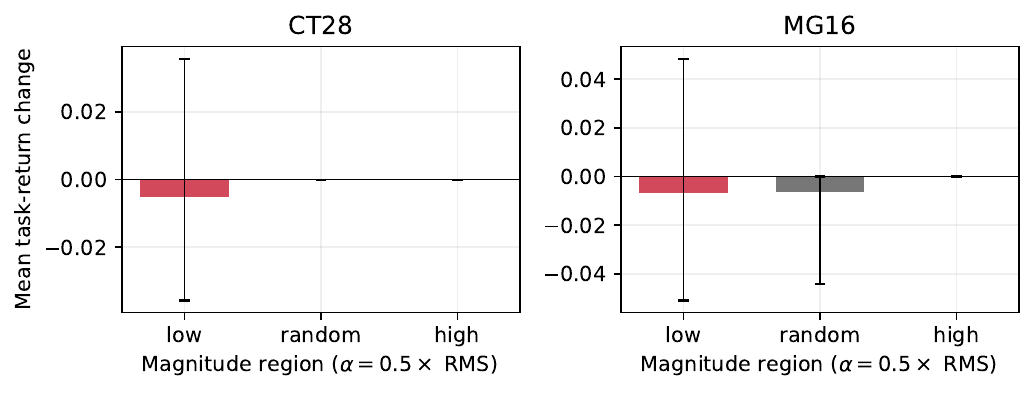}
        \textbf{(d) Diagnostic return change at \(\alpha=0.5\).}
    \end{minipage}
    \caption{Mechanistic analysis of continuous mask scores. Panel (a) gives
    the empirical cumulative distribution of up to \(2{,}048\) sampled scores
    per task--layer tensor, normalized by layer RMS. Panels (b)--(d) perturb
    \(25\%\) of scores from each magnitude region. Lines report seed means;
    bands in (b) and (c) span seed minima and maxima. Return estimates in (d)
    use three episodes per task.}
    \label{fig:mask_score_perturbation}
\end{figure*}

\paragraph{Perturbation results.}
Across actor heads and shared trunks, \(3.7\)--\(4.8\%\) of scores lie below
\(0.05\) RMS and \(7.1\)--\(9.4\%\) lie below \(0.10\) RMS. At
\(\alpha=0.5\), perturbing the lowest-magnitude quartile changes \(37.3\%\) of
the selected CT28 gates and \(39.3\%\) of the selected MG16 gates. The
corresponding rates are \(14.2\%\) and \(16.4\%\) for random scores, but only
\(0.2\%\) and \(0.5\%\) for the highest-magnitude quartile. Low-magnitude
perturbations also change the selected action on \(6.5\%\) of CT28 states and
\(6.2\%\) of MG16 states, compared with \(0.8\%\) for high-magnitude
perturbations on both benchmarks. One-sided Wilcoxon signed-rank tests support
the paired low-versus-high differences in CT28 (\(p=0.0078\)) and MG16
(\(p=0.017\)).

Mean task-return changes at \(\alpha=0.5\) are \(-0.005\) on CT28 and
\(-0.007\) on MG16 for low-magnitude perturbations. The low-versus-high return
differences are not significant (\(p=0.60\) and \(p=0.61\), respectively).
Score magnitude therefore predicts gate and action stability, but the
diagnostic rollout data do not establish that low-magnitude scores can be
removed without a performance cost. The difference between action changes and
mean-return changes also indicates behavioral redundancy in the learned
subnetworks.

\paragraph{Quantized mask libraries.}
The second experiment tests whether continuous completed-task scores are
required for subsequent composition. Binary storage retains only score sign,
\begin{equation}
    Q_{\mathrm{bin}}(s)=
    \begin{cases}
        +1,&s>0,\\
        -1,&s\leq0,
    \end{cases}
    \label{eq:binary_score_quantization}
\end{equation}
whereas ternary storage introduces an RMS-scaled dead zone,
\begin{equation}
    Q_{\delta}(s;r)=
    \begin{cases}
        +1,&s>\delta r,\\
        0,&|s|\leq\delta r,\\
        -1,&s<-\delta r.
    \end{cases}
    \label{eq:ternary_score_quantization}
\end{equation}
Increasing \(\delta\) widens the dead zone and maps a larger fraction of the
library to zero.
The value \(0\) removes a near-boundary connection. The value \(-1\) remains
inactive when a stored task is evaluated alone. These values differ during
composition because AMSC combines score tensors before thresholding. For a
new task \(k\), the composed score is
\begin{equation}
    \widehat S_k^{\ell}
    =a_{k,k}\mathcal N(S_k^{\ell})
    +\sum_{i\in\sP(k)}a_{k,i}
        \mathcal N(Q(S_i^{\ell})),
    \qquad
    M_k^{\ell}=\mathbb I[\widehat S_k^{\ell}>0],
    \label{eq:quantized_mask_composition}
\end{equation}
where \(\mathcal N\) is the L2-STE transformation and \(Q\) is either
Equation~\ref{eq:binary_score_quantization} or
Equation~\ref{eq:ternary_score_quantization}. After task \(k\), its
consolidated score is quantized before entering the reusable library. The
experiment therefore evaluates online learning with quantized sources rather
than post-hoc compression of an already completed run.

\begin{figure*}[t]
    \centering
    \begin{minipage}{0.86\textwidth}
        \centering
        \includegraphics[width=\linewidth]{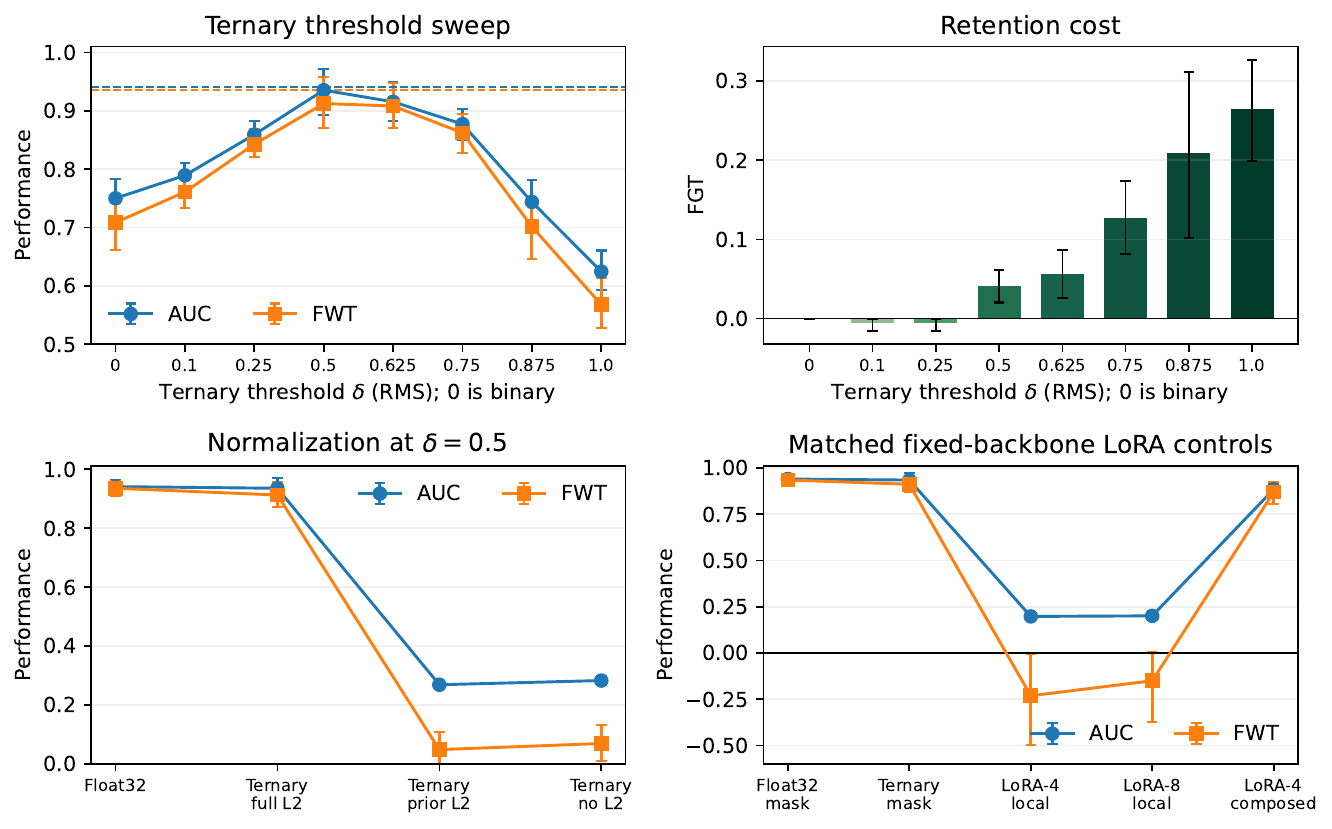}
        \textbf{(a) Quantization, normalization, and matched LoRA controls.}
    \end{minipage}

    \vspace{0.7em}
    \begin{minipage}{0.86\textwidth}
        \centering
        \includegraphics[width=\linewidth]{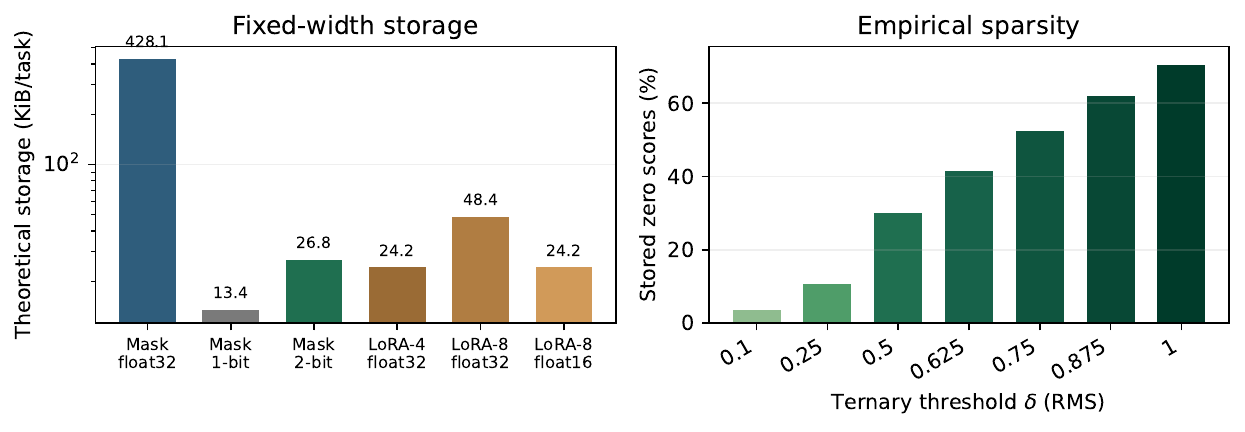}
        \textbf{(b) Theoretical storage and ternary sparsity.}
    \end{minipage}
    \caption{Quantized storage of completed task masks on CT28. In (a), dashed
    lines in the threshold sweep give Float32-mask performance and error bars
    denote asymmetric \(95\%\) bootstrap confidence intervals across seven
    seeds. The LoRA controls use the same frozen signed-constant backbone.
    Storage in (b) counts packed values only and excludes container and index
    overhead.}
    \label{fig:mask_score_quantization}
\end{figure*}

\paragraph{Quantization results.}
Float32 AMSC obtains AUC/FWT/FGT of \(0.941/0.935/0.000\). Binary storage
reduces AUC/FWT to \(0.750/0.709\) despite preserving each stored score's
sign. Score sign alone therefore does not preserve the information required
by later composition. Ternary storage reaches its best trade-off at
\(\delta=0.50\), obtaining \(0.935/0.912/0.041\) while setting \(29.9\%\) of
stored scores to zero. Paired tests against Float32 do not reject zero AUC
(\(p=0.43\)) or FWT difference (\(p=0.22\)), although this does not establish
equivalence. FGT increases by \(0.041\) (\(p=0.031\)). At
\(\delta=0.625\), AUC/FWT/FGT is \(0.915/0.908/0.056\) with \(41.4\%\) zeros;
the AUC reduction is significant (\(p=0.023\)) while the FWT reduction is not
(\(p=0.086\)). Increasing \(\delta\) further monotonically increases sparsity
but reduces acquisition and retention, reaching \(0.624/0.568/0.265\) at
\(\delta=1\). Thus, the dead zone can remove near-boundary connections, but
aggressive thresholding deletes gates required by completed policies.

\paragraph{Why normalization is necessary.}
For a ternary tensor, \(\lVert Q_\delta(S_i^\ell)\rVert_F=\sqrt{n_i^\ell}\),
where \(n_i^\ell\) is its number of nonzero entries. Without normalization,
the effective influence of source \(i\) is therefore proportional to
\(a_{k,i}\sqrt{n_i^\ell}\), rather than the intended similarity coefficient
\(a_{k,i}\). The trainable current score introduces an additional arbitrary
continuous scale. L2-STE gives every component unit forward norm, making the
composition coefficients comparable while retaining the identity gradient
for the current score. This calibration is particularly important after
ternary quantization because all retained entries have magnitude one, leaving
density as the only source-dependent tensor scale. At \(\delta=0.50\), removing
normalization yields AUC/FWT/FGT of \(0.283/0.070/0.291\). Normalizing only the
priors also fails (\(0.269/0.049/0.005\)) because unit-norm sources remain
incommensurate with the raw current score. Joint normalization is therefore
required for sparsemax weights to control actual mask contributions.

\paragraph{Storage comparison.}
CT28 contains \(D_m=109{,}600\) mask scores per task. Float32 scores require
\(428.1\) KiB per task. Packed binary and ternary scores require \(13.4\) and
\(26.8\) KiB, respectively. For comparison, a rank-\(r\) LoRA adapter
\citep{hu2022lora} over the same layers stores
\begin{equation}
    D_{\mathrm{LoRA}}(r)
    =r\sum_{\ell}(d_{\ell}^{\mathrm{in}}+d_{\ell}^{\mathrm{out}})
    =1{,}548r
    \label{eq:lora_storage_reference}
\end{equation}
values per task. Rank-4 and rank-8 Float32 adapters therefore require
\(24.2\) and \(48.4\) KiB; rank-8 Float16 also requires \(24.2\) KiB. Ternary masks are similar in theoretical size to rank-4 LoRA and smaller than rank-8 Float32 LoRA, whereas binary masks are smaller than both. A task-local rank-4 adapter obtains AUC/FWT/FGT of
\(0.198/-0.231/0.000\), and rank 8 obtains
\(0.201/-0.150/0.000\). Both fall below the \(0.22\) AUC of full
from-scratch learning. They combine a low-rank capacity constraint with no
mechanism for reusing earlier solutions, which is especially restrictive in
the hierarchical CT28 curriculum. In contrast, normalized-sparsemax retrieval
and composition of rank-4 deltas raises performance to
\(0.885/0.870/0.000\). At \(24.2\) KiB per task, this adapter is slightly
smaller than a packed ternary mask, but remains below the latter by \(0.050\)
AUC (paired \(95\%\) CI \([0.020,0.087]\)) and \(0.042\) FWT
(\([-0.016,0.101]\)). Composed LoRA must express and recompress the transferred
update within a rank-4 subspace, whereas mask composition can independently
combine all frozen backbone connections. The LoRA controls retain zero
measured forgetting, while ternary quantization incurs a small boundary loss
of \(0.041\) (\([0.020,0.061]\)) when a completed continuous score is replaced
by its quantized representation.
These results suggest that quantized masks can approach low-rank-adapter storage
while preserving stronger composition in this setting.


\section{Implementation Details}
\label{sec:implementation_details}

\subsection{Wasserstein Task Embedding Construction}
\label{sec:embedding_details}
AMSC constructs WTEs from experience generated by the current policy
\citep{liu2022wasserstein}. The current-task buffer \(\sD_k\) stores
transitions \((\vs,a,r,\vs',d)\). At an embedding update, AMSC samples \(N\) \((\vs,a,r)\) entries uniformly with replacement from \(\sD_k\). The stored state has already received
the benchmark-specific policy-input scaling. Let \(d_s\) denote the flattened
state dimension and let \(d_a\) denote the action-feature dimension. Discrete
actions are one-hot encoded; continuous actions retain their vector
representation:
\begin{equation}
    \phi(a_n)=
    \begin{cases}
        \operatorname{onehot}(a_n)\in\R^{d_a},
            & \text{discrete actions},\\
        a_n\in\R^{d_a},
            & \text{continuous actions}.
    \end{cases}
    \label{eq:wte_action_encoding}
\end{equation}

The state block is standardized independently for each sampled batch. The
implementation computes one scalar mean and standard deviation over all
\(N d_s\) state entries, rather than separate statistics for each state
coordinate. For batch update \(t\),
\begin{equation}
    \mu_s^{(t)}=\operatorname{mean}_{n,j}s_{n,j},\qquad
    \sigma_s^{(t)}=\operatorname{std}_{n,j}s_{n,j},\qquad
    \widehat{\vs}_n=
    \frac{\vs_n-\mu_s^{(t)}}{\max(\sigma_s^{(t)},\varepsilon_x)}.
    \label{eq:wte_state_preprocessing}
\end{equation}
The implementation uses \(\varepsilon_x=10^{-8}\).
The input row and embedding batch are then
\begin{equation}
    \vx_n=[\widehat{\vs}_n^\top,\phi(a_n)^\top,r_n]^\top,
    \qquad
    \mX_k^{(t)}=[\vx_1,\ldots,\vx_N]^\top
        \in\R^{N\times d_x},
    \label{eq:wte_input_batch}
\end{equation}
where \(d_x=d_s+d_a+1\). The reward is used at the scale stored by the RL
replay buffer; WTE applies no additional reward normalization.

At initialization, AMSC samples a reference matrix
\(\mX_0=[\vx_1^0,\ldots,\vx_R^0]^\top\in\R^{R\times d_x}\), with each
coordinate sampled independently from \(\mathcal{U}[0,1]\). A seeded random
generator draws \(\mX_0\) once, and the same ordered reference is retained for
all tasks in a run. The reference is not estimated from a task and is not
updated during training. Reusing \(\mX_0\) fixes the coordinates and dimension
of every task embedding.

The empirical batch and reference distributions have uniform masses
\begin{equation}
    \va=\frac{1}{N}\vone_N,\qquad
    \vb=\frac{1}{R}\vone_R.
    \label{eq:wte_marginals}
\end{equation}
Let \(C_{n,r}=\lVert\vx_n-\vx_r^0\rVert_2^2\) be entry \((n,r)\) of the
cost matrix \(\mC\). AMSC uses the unregularized
Earth Mover's Distance solver in the Python Optimal Transport library to
compute
\begin{equation}
    \bm{\Gamma}^\star
    =\underset{\bm{\Gamma}\in\Pi(\va,\vb)}
        {\operatorname{argmin}}
        \langle\bm{\Gamma},\mC\rangle_F,
    \quad
    \Pi(\va,\vb)
    =\left\{\bm{\Gamma}\in\R_+^{N\times R}:
    \bm{\Gamma}\vone_R=\va,\,
    \bm{\Gamma}^\top\vone_N=\vb\right\}.
    \label{eq:wte_transport_plan}
\end{equation}
Since each reference point receives mass \(1/R\), its barycentric projection
into the current empirical distribution is
\begin{equation}
    \vy_r=R\sum_{n=1}^{N}\Gamma^\star_{n,r}\vx_n.
    \label{eq:wte_barycentric_projection}
\end{equation}
The WTE records the displacement of every ordered reference point under this
projection:
\begin{equation}
    \ve_k^{(t)}
    =\frac{1}{\sqrt{R}}
      \operatorname{vec}\!\left(
      [\vy_1-\vx_1^0,\ldots,\vy_R-\vx_R^0]^\top\right)
    \in\R^{R d_x}.
    \label{eq:wte_embedding_details}
\end{equation}
Equation~\eqref{eq:wte_embedding_details} is the expanded form of
Equation~\eqref{eq:amsc_WTE}. The construction contains no learned encoder or
embedding loss.

The first update initializes
\(\bar{\ve}_k^{(1)}=\operatorname{normalize}(\ve_k^{(1)})\).
Subsequent updates combine the stored normalized representation with the raw
batch embedding, then normalize the result,
\begin{equation}
    \bar{\ve}_k^{(t)}
    =
    \operatorname{normalize}\!\left(
        \rho\bar{\ve}_k^{(t-1)}+(1-\rho)\ve_k^{(t)}
    \right).
    \label{eq:wte_online_update_details}
\end{equation}
Here,
\(\operatorname{normalize}(\ve)=
\ve/\max(\lVert\ve\rVert_2,\varepsilon_e)\),
with \(\varepsilon_e=10^{-8}\) and \(\rho=0.5\).
This is the update in Equation~\eqref{eq:amsc_ema} used for the reported
CT28 and MG16 experiments. Completed embeddings are frozen at task boundaries, and the current-task buffer is cleared before the next task. Every retrieval event therefore
compares an evolving current-task estimate with fixed representations of the
completed tasks.

The reported CT28 and MG16 experiments use \(R=50\). CT28 samples \(N=128\)
transitions and updates both the embedding and retrieval support every PPO
iteration. MG16 samples \(N=1024\), updates the embedding every iteration, and
updates the retrieval support every five iterations. The CW10 implementation
uses \(R=50\), \(N=1000\), and an update period of \(1000\) environment steps.

\subsection{Pseudo-code}
\label{sec:pseudocode}
Algorithm~\ref{alg:algorithm} gives the AMSC training procedure. Task
boundaries determine when a completed mask and embedding are frozen and added
to their respective libraries. Embedding and retrieval updates use periods
\(f_{\mathrm{emb}}\) and \(f_{\mathrm{sel}}\), respectively. Retrieval
weights are not optimized by the RL loss; gradients update only the current
task's mask scores.

\begin{algorithm}[h]
    \caption{AMSC training with adaptive similarity-guided selection}
    \label{alg:algorithm}
    \small
    \begin{algorithmic}[1]
        \Require Task sequence \(\{\mathcal{T}_k\}_{k=1}^{K}\), per-task training budget \(B\), rollout length \(z\)
        \Require WTE batch size \(N\), reference size \(R\), EMA factor \(\rho\), sparsemax temperature \(\tau\)
        \Require Update periods \(f_{\mathrm{emb}}\) and \(f_{\mathrm{sel}}\)
        \State Initialize frozen backbone \(\{\mW^\ell\}_{\ell=1}^{L}\), mask library \(\sM\), and embedding table \(\mE\)
        \State Sample and fix reference matrix \(\mX_0\in\R^{R\times d_x}\)
        \For{\(k=1\) \textbf{to} \(K\)}
            \State Initialize current mask-score matrices \(\{\mS_k^\ell\}_{\ell=1}^{L}\)
            \State Set \(\sD_k\gets\varnothing\), \(\sP(k)\gets\varnothing\), \(steps\gets0\), and \(t\gets0\)
            \While{\(steps<B\)}
                \State Form \(\{\mC_k^\ell\}_{\ell=1}^{L}\) and the binary policy masks using Equation~\eqref{eq:amsc_composition}
                \State Collect a rollout of \(z\) transitions using the resulting policy
                \State Append transitions to \(\sD_k\); update \(steps\gets steps+z\) and \(t\gets t+1\)
                \State Compute the RL loss and update only \(\{\mS_k^\ell\}_{\ell=1}^{L}\) through Equation~\eqref{eq:amsc_l2ste}
                \If{\(|\sD_k|\geq N\) and \(t\bmod f_{\mathrm{emb}}=0\)}
                    \State Sample \(N\) tuples from \(\sD_k\) and construct \(\mX_k^{(t)}\)
                    \State Compute \(\ve_k^{(t)}\) using Equations~\eqref{eq:wte_transport_plan}--\eqref{eq:wte_embedding_details}
                    \If{this is the first embedding update for task \(k\)}
                        \State Set \(\bar{\ve}_k\gets\operatorname{normalize}(\ve_k^{(t)})\)
                    \Else
                        \State Update \(\bar{\ve}_k\) using Equation~\eqref{eq:amsc_ema}
                    \EndIf
                \EndIf
                \If{\(k>1\), \(\bar{\ve}_k\) exists, and \(t\bmod f_{\mathrm{sel}}=0\)}
                    \State Compute \(\vs_k\), \(\vw_k\), and \(\sP(k)\) using Equations~\eqref{eq:amsc_similarity}--\eqref{eq:amsc_retrieval}
                \EndIf
            \EndWhile
            \State Recompute \(\{\mC_k^\ell\}_{\ell=1}^{L}\) using Equation~\eqref{eq:amsc_composition}
            \State Store \(\mS_k^{\ell,*}\gets\mC_k^\ell\) for every \(\ell\) in \(\sM\), and store \(\bar{\ve}_k\) in \(\mE\)
        \EndFor
    \end{algorithmic}
\end{algorithm}

\section{Benchmarks}
\label{sec:benchmarks}
For CT28 and MG16, we use \emph{family} to denote tasks that share an
observation encoding and a nested task construction across depths. All tasks
within each benchmark have the same formal observation dimensions and action
interface. Cross-family tasks differ in their observation dataset or symbolic
remapping, rather than in tensor shape, and can still share behavioral
structure that supports policy transfer.

\paragraph{CT-Graph CT28.}
CT-Graph is a configurable decision-tree environment for continual RL \citep{soltoggio2019ctgraph}. At each decision state, the agent selects a branch and receives a sparse reward of \(1\) only after reaching the target terminal node. Each state is represented by an image generated from a family-specific image dataset. Datasets do not overlap across families, whereas tasks within one family share observations for their common graph prefix.

CT28 is based on the evaluation setup used in \citep{nath2025policysearchretrievalcomposition}, and contains four families with depths 2 through 8, giving 28 tasks. Within each family, the solution at depth \(d\) forms a prefix of the solution at depth \(d+1\). A depth-8 target is encountered after 24--27 previous tasks. Six of the available policies are lower-depth members of the target family, and the remaining policies belong to other families. CT28 therefore tests retrieval from a library in which relevant predecessors form a minority that decreases as the curriculum progresses.

\paragraph{MiniGrid MG16.}
MiniGrid provides partially observable grid-world navigation tasks with symbolic object, color, and state observations \citep{gym_minigrid}. MG16 extends MultiRoom into four room-chain families. Each family contains tasks R2 through R5. Under a fixed layout seed, R\(d+1\) preserves the complete R\(d\) layout and adds one room and connecting transition. The agent receives sparse reward for reaching the goal in the final room. The action space and physical task remain fixed across families.

\begin{figure}[h]
    \centering
    \includegraphics[width=1.0\textwidth]{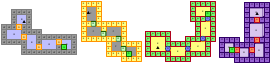}
    \caption{Depth-5 layouts for the four MG16 task families. Symbols denote the observed representation of each functional cell; the underlying wall, door, and floor affordances remain unchanged. Identity uses grey walls (\texttt{W}), blue ball floors (\texttt{B}), and yellow doors (\texttt{D}). KeyBoxSwap uses yellow key walls (\texttt{K}), grey floors (\texttt{F}), and green box doors (\texttt{B}). LavaBall uses green lava walls (\texttt{L}), yellow key floors (\texttt{K}), and blue ball doors (\texttt{O}). BallLava uses purple ball walls (\texttt{O}), purple box floors (\texttt{B}), and red lava doors (\texttt{L}).}
    \label{fig:minigrid-benchmark}
\end{figure}

MG16 assigns each family a distinct symbolic remapping of wall, door, and floor cells. The remapping changes the observed object type and colour but preserves collision, traversal, and door-toggle affordances. Each remapping remains fixed across depth, so lower-depth policies share the target family's observation code. Figure~\ref{fig:minigrid-benchmark} shows the four profiles. Tasks are ordered by room depth, giving four R2 tasks, followed by four R3, R4, and R5 tasks. No family identifier is included in the policy observation.

\paragraph{Continual World CW10.}
Continual World evaluates continual RL on sequences of MetaWorld robotic
manipulation tasks \citep{wolczyk2021continualworld}. We use one pass through
the ten-task CW10 curriculum: \texttt{hammer}, \texttt{push-wall},
\texttt{faucet-close}, \texttt{push-back}, \texttt{stick-pull},
\texttt{handle-press-side}, \texttt{push}, \texttt{shelf-place},
\texttt{window-close}, and \texttt{peg-unplug-side}. Each task is trained for
\(1\mathrm{M}\) environment steps with SAC, giving \(10\mathrm{M}\) steps in
total. CW10 provides shaped rewards and four-dimensional continuous actions.
It therefore tests AMSC beyond the sparse-reward, discrete-action conditions
of CT28 and MG16.

Our experiments use the current MetaWorld-v2 goal-observable task
definitions \citep{yu2020meta}. Each observation is a 39-dimensional state
vector, and no one-hot task label is appended. Initial object and goal
configurations remain randomized across episodes. 

Evaluation uses binary task success rather than shaped return. An episode is
successful if the environment reports success at any step before termination.
Every \(50{,}000\) training steps, the deterministic policy is evaluated for
ten episodes on every task encountered so far. Each evaluation uses a fresh
environment and a task-specific seed, and does not alter the training
environment or replay buffer. The resulting success-rate matrix supplies the
CW10 AUC, FWT, and FGT measurements.
\begin{figure}[h]
    \centering
    \includegraphics[width=0.9\textwidth]{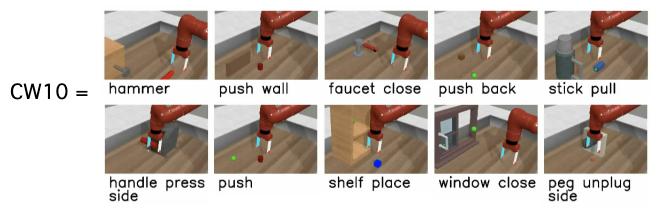}
    \caption{The ten MetaWorld-v2 tasks in the CW10 curriculum, shown in
    execution order from left to right and top to bottom \citet{wolczyk2021continualworld}.}
    \label{fig:continualworld-benchmark}
\end{figure}

\section{Baselines}
\label{sec:baselines_extended}

\paragraph{Comparison protocol.}
All methods use seven seeds, identical task orders, and the same interaction
budget. CT28 and MG16 use the common PPO architecture and optimizer settings
reported in Section~\ref{sec:results}. Task labels are used at task switches
and evaluation but are not included in policy observations or policy selection. CW10 uses the SAC schedule and evaluation protocol in Appendix~\ref{sec:benchmarks}.
Architectures are matched where the algorithms permit direct substitution.
CompoNet and ProgNet retain their released two-layer, 256-unit ReLU models and a
matching no-reuse run is used to calculate their FWT. Other CW10 methods use the
four-layer, 256-unit CW architecture with first-layer LayerNorm.

\paragraph{Learning references and regularization.}
Fine-tuning updates one policy throughout the curriculum without a retention
mechanism. No-reuse trains an independent policy on each task and provides the
zero-transfer learning curves used to calculate FWT. 

\paragraph{Replay.}
SER \citep{isele2018selective} maintains short-term FIFO memory and bounded episodic memory. We report surprise-prioritized and reservoir-sampled SER variants. SDW
\citep{huang2025lifelongreinforcementlearningsimilaritydriven} estimates task
similarity and uses it to control replay and policy/value consistency weights.
These methods select or weight stored experience rather than stored policy
modules.

\paragraph{Parameter isolation and routing.}
ProgNet \citep{rusu2016progressive} adds a frozen policy column for each task and learns lateral connections from earlier columns. OWL \citep{kessler2022state} combines task-specific heads with oEWC on shared features and selects a head using an exponential-weights bandit. SANE \citep{powers2022neuralensembles} maintains an ensemble of actor--critic
modules and activates a module using optimistic value uncertainty. Our
task-incremental SANE configuration clones the selected module at each task
boundary and freezes earlier modules.

\paragraph{Policy composition.}
Mask-LC and Mask-BLC \citep{ben2022lifelong} learn softmax coefficients over
the current mask and every earlier mask. Mask-LC initializes the active
coefficients uniformly. Mask-BLC initially allocates half of the mass to the
current mask and divides the remainder among the earlier masks. CKA-RL
\citep{hu2025continual} learns task-specific parameter residuals around a
shared policy and fuses residuals into a bounded knowledge pool. CompoNet
\citep{malagon2024composing} adds one policy unit per task and uses learned
attention to combine the outputs of frozen earlier units with the current
unit. These baselines learn routing or composition without first constructing
a library-relative sparse support.

\paragraph{Similarity-guided composition.}
MOSAIC \citep{nath2025policysearchretrievalcomposition} was introduced for
knowledge exchange among concurrently learning agents. We adapt MOSAIC as a baseline, limiting candidates to frozen masks from earlier tasks but retaining MOSAIC's absolute cosine threshold, performance gate, learned composition, and acquisition protocol. A successful acquisition consolidates the active composition and reinitializes its coefficients from source and target performance. We also evaluate MOSAIC without the
performance gate, denoted MOSAIC \(\neg\mathbb{I}_{\mathrm{perf}}\). 

\section{Hyperparameters}
\label{sec:hyperparameters}
Table~\ref{tbl:hyperparameters_paper3} reports the principal AMSC
configuration and shared optimization settings.
Table~\ref{tbl:hyperparameters_modular} reports method-specific settings for the
CT28 and MG16 baselines. Table~\ref{tbl:hyperparameters_cw10} reports the
corresponding CW10 settings. Methods share the optimization and evaluation
protocol for each benchmark unless stated otherwise.
\begin{table}[h]
    \caption{Hyperparameters, seeds, and training budgets for the principal
    AMSC experiments. One CT28 or MG16 PPO iteration collects
    \(4\times128=512\) environment samples. The CW10 configuration uses
    the 39-dimensional MetaWorld-v2 observations.}
    \label{tbl:hyperparameters_paper3}
    \begin{center}
    \small
    \renewcommand{\arraystretch}{1.02}
    \begin{tabularx}{\textwidth}{llXXX}
        \toprule
        \textbf{Component} & \textbf{Hyperparameter}
        & \textbf{CT28} & \textbf{MG16} & \textbf{CW10} \\
        \midrule
        Optimization
            & Learner & PPO & PPO & SAC \\
            & Network width
            & \(3\times200\) & \(3\times200\) & \(4\times256\) \\
            & Activation
            & ReLU & ReLU & Tanh (first), LeakyReLU (remaining) \\
            & Optimizer
            & RMSprop & RMSprop & Adam \\
            & Actor/critic learning rate
            & \(1.5\times10^{-4}\) & \(1.5\times10^{-4}\)
            & \(10^{-3}/10^{-3}\) \\
            & Number of workers & 4 & 4 & 1 \\
            & Discount factor & 0.99 & 0.99 & 0.99 \\
            & GAE parameter & 0.99 & 0.99 & --- \\
            & Entropy coefficient & 0.1 & 0.01 & Automatic \\
            & Rollout length per worker & 128 & 128 & --- \\
            & Optimization epochs & 8 & 8 & --- \\
            & Number of mini-batches & 64 & 64 & --- \\
            & PPO ratio clip & 0.1 & 0.1 & --- \\
            & Gradient clipping & 5 & 5 & --- \\
            & Replay-buffer size & --- & --- & \(10^6\) \\
            & Replay batch size & --- & --- & 128 \\
            & Learning starts & --- & --- & \(1{,}000\) steps \\
            & Initial random actions & --- & --- & \(10{,}000\) steps \\
            & Update interval & --- & --- & 50 steps \\
            & Gradient steps per update & --- & --- & 50 \\
            & Target smoothing coefficient & --- & --- & 0.005 \\
        \midrule
        AMSC
            & Task representation & WTE--SAR & WTE--SAR & WTE--SAR \\
            & Reference size \(R\) & 50 & 50 & 50 \\
            & Embedding sample count \(N\) & 128 & 1024 & 1,000 \\
            & Embedding-update frequency
            & 1 iteration & 1 iteration & \(1{,}000\) steps \\
            & Selection frequency
            & 1 iteration & 5 iterations & \(1{,}000\) steps \\
            & EMA coefficient \(\rho\) & 0.5 & 0.5 & 0.5 \\
            & Similarity normalization & z-score & z-score & z-score \\
            & Sparsemax temperature \(\tau\) & 1.0 & 1.0 & 1.0 \\
            & Mask-score normalization
            & \(\ell_2\)-STE & \(\ell_2\)-STE & \(\ell_2\)-STE \\
            & Mask type and scope
            & Discrete actor--critic & Discrete actor--critic
            & Discrete full actor \\
            & Critic & Masked & Masked & Dense, task-local \\
            & LayerNorm & None & None & Task-specific affine \\
        \midrule
        Training
            & Budget per task
            & \(51{,}200\) steps & \(512{,}000\) steps
            & \(1{,}000{,}000\) steps \\
            & Total budget
            & \(1{,}433{,}600\) steps & \(8{,}192{,}000\) steps
            & \(10{,}000{,}000\) steps \\
            & Evaluation interval
            & 10 iterations & 20 iterations & \(50{,}000\) steps \\
            & Evaluation episodes & 10 & 10 & 10 per seen task \\
        \midrule
        Seeds
            & Experiment seeds
            & \multicolumn{3}{c}{\(86,87,88,89,90,91,92\)} \\
            & CT-Graph configuration seed & 3 & --- & --- \\
            & CT-Graph image seeds & \(1,2,3,4\) & --- & --- \\
        \bottomrule
    \end{tabularx}
    \end{center}
\end{table}
\begin{table}[h]
    \caption{Method-specific CT28 and MG16 hyperparameters for parameter
    isolation, routing, and policy-composition methods.}
    \label{tbl:hyperparameters_modular}
    \begin{center}
    \small
    \renewcommand{\arraystretch}{0.98}
    \begin{tabularx}{\textwidth}{l|XX}
    \toprule
    \textbf{Algorithm} & \textbf{Hyperparameter}
        & \textbf{Value (CT28/MG16)} \\
    \midrule
    SER
        & Storage strategy & Surprise/reservoir \\
        & Episodic-memory size & 20,000 \\
        & FIFO-memory size & 2,000 \\
        & Batch size & 256 \\
        & Replay epochs & 1 \\
        & FIFO ratio & 0.5 \\
        & Coverage distance & 1.0 \\
    \midrule
    SDW
        & Buffer size & \(10^5\) \\
        & Batch size & 256 \\
        & Probe batch size & 128 \\
        & Base replay ratio & 0.8 \\
        & Similarity smoothing & 2 \\
        & Maximum policy/value weights & \(0.01/0.005\) \\
        & Minimum consistency weight & \(-10^{-4}\) \\
        & Similarity-feature weights & \(0.4,0.4,0.2\) \\
    \midrule
    OWL
        & oEWC coefficient & 500 \\
        & Bandit learning rate & 0.88 \\
        & Bandit decay & 0.9 \\
        & Exploration \(\epsilon\) & 0 \\
    \midrule
    SANE
        & Maximum modules & CT28: 28; MG16: 16 \\
        & Uncertainty scale & 1.0 \\
        & Task-boundary cloning & Enabled \\
    \midrule
    Mask-LC
        & Mask type & Discrete \\
        & Prior support & Full library \\
        & Coefficient initialization & Uniform \\
    \midrule
    Mask-BLC
        & Mask type and support & Same as Mask-LC \\
        & Initial composition mass & Current: 0.5; all priors: 0.5 \\
    \midrule
    CKA-RL
        & Knowledge-pool capacity & 5 \\
        & \(\alpha\) scale & \(10^{-2}\) \\
        & \(\alpha\) learning rate & \(2.5\times10^{-4}\) \\
        & \(\alpha\) initialization & Random normal \\
        & Major mass & 0.6 \\
        & Fixed/global/scaled \(\alpha\) & Fixed: no; global: yes; scaled: yes \\
        & Fused components & Actor: yes; encoder: no \\
        & \(\Delta\Theta\) mode & T \\
        & Reset critic & Enabled \\
    \midrule
    CompoNet
        & Attention dimension & 200 \\
        & Composed output & Policy logits \\
    \midrule
    MOSAIC-LL
        & Task representation & WTE--SAR \\
        & Reference size \(R\) & 50 \\
        & Embedding sample count \(N\) & 128 \\
        & Cosine-similarity threshold & 0.5 \\
        & Selection cap & None \\
        & Selection frequency & CT28: 10; MG16: 25 iterations \\
        & Performance gate & Source return \(>\) current target return \\
        & Acquisition update & Consolidate composition and reset coefficients \\
        & Coefficient initialization & Reward-guided \\
    \midrule
    MOSAIC \(\neg\mathbb{I}_{\mathrm{perf}}\)
        & Performance gate & Disabled; otherwise as MOSAIC-LL \\
    \bottomrule
    \end{tabularx}
    \end{center}
\end{table}
\begin{table}
    \caption{Method-specific CW10 baseline hyperparameters. Compatible methods
    use the shared SAC configuration in
    Table~\ref{tbl:hyperparameters_paper3}. CompoNet and ProgNet retain their
    released architectures and use the matched native no-reuse run as their
    FWT reference.}
    \label{tbl:hyperparameters_cw10}
    \begin{center}
    \small
    \renewcommand{\arraystretch}{0.98}
    \begin{tabularx}{\textwidth}{l|XX}
    \toprule
    \textbf{Algorithm} & \textbf{Hyperparameter} & \textbf{CW10 value} \\
    \midrule
    No-reuse
        & Network & \(4\times256\), first-layer LayerNorm \\
    \midrule
    No-reuse (for CompoNet and ProgNet)
        & Network & \(2\times256\), ReLU, no LayerNorm \\
    \midrule
    Mask-LC
        & Mask type & Continuous \\
        & Mask scope & Actor and critics, trunks and heads \\
        & LayerNorm & Task-specific affine \\
        & Prior support & Full library \\
        & Composition & Learned \\
        & Coefficient initialization & Uniform \\
    \midrule
    Mask-BLC
        & Mask type, scope, and support & Same as Mask-LC \\
        & Initial composition mass & Current: 0.5; all priors: 0.5 \\
    \midrule
    CKA-RL
        & Knowledge-pool capacity & 8 \\
        & \(\alpha\) initialization & Random normal \\
        & \(\alpha\) scale & \(10^{-3}\) \\
        & Major mass & 0.6 \\
        & Fixed/global/scaled \(\alpha\) & Fixed: no; global: yes; scaled: yes \\
        & Fused components & Heads: yes; shared trunk: no \\
        & \(\Delta\Theta\) mode & T \\
    \midrule
    CompoNet
        & Network & \(2\times256\), ReLU, no LayerNorm \\
        & Composition & Learned attention \\
    \midrule
    ProgNet
        & Network & \(2\times256\), ReLU, no LayerNorm \\
        & Prior columns & Frozen \\
    \midrule
    MOSAIC-LL
        & Mask type and scope & Continuous full network \\
        & LayerNorm & Task-specific affine \\
        & Task representation & WTE--SAR \\
        & Reference size \(R\) & 50 \\
        & Embedding sample count \(N\) & 1,000 \\
        & Selection frequency & \(1{,}000\) steps \\
        & Cosine-similarity threshold & 0.5 \\
        & Performance estimate & Mean success over 10 recent online episodes \\
        & Performance gate & Source success \(>\) current target success \\
        & Acquisition update & Consolidate composition and reset coefficients \\
        & Coefficient initialization & Success-guided \\
    \bottomrule
    \end{tabularx}
    \end{center}
\end{table}

\section{Additional Results}
\label{sec:additional_results}

\subsection{Early Acquisition}
\label{sec:early_acquisition}
We compare early acquisition against MOSAIC, which also retrieves masks for
composition, and the surprise- and reservoir-based SER replay methods.
\(\mathrm{AUC}_{20}\) averages the first two of ten target-task evaluations,
covering the first 20\% of each CT28 task budget. Time to first reward is
the end of the first 512-step training iteration with positive mean reward;
time to first positive evaluation return is measured at evaluation
checkpoints. Both times are right-censored at the \(51{,}200\)-step task
budget. Aggregate statistics exclude task 0 because no source policy is
then available.

Across tasks 1--27, AMSC obtains \(\mathrm{AUC}_{20}=0.86\), compared
with \(0.35\) for MOSAIC (two-sided paired \(p=0.016\)). Its first positive
training reward appears after \(2.34\mathrm{k}\) rather than
\(5.83\mathrm{k}\) steps (\(p=0.016\)). SER-surprise achieves
\(\mathrm{AUC}_{20}=0.91\) and reaches positive reward after
\(1.59\mathrm{k}\) steps; SER-reservoir obtains \(0.86\) and
\(3.27\mathrm{k}\). Neither SER variant differs significantly from AMSC
on early AUC or time to first reward. The MOSAIC comparison supports an
early advantage for AMSC within mask-based composition; it does not imply
an early advantage over replay. Tables~\ref{tbl:ct28_early_acquisition_summary}
and~\ref{tbl:ct28_early_acquisition} provide the aggregate and per-task
results.

\begin{table}[h]
    \caption{CT28 early acquisition over tasks 1--27 across seven matched seeds. Entries are means with conservative 95\% bootstrap CI. $T_{+}$ and $T_{\mathrm{eval}+}$ are time to first positive training reward and evaluation return, in thousands of environment steps, right-censored at $51.20$k. Exact two-sided paired sign-flip $p$-values compare AMSC with each method; no multiple-comparison adjustment is applied.}
    \label{tbl:ct28_early_acquisition_summary}
    \centering
    \scriptsize
    \setlength{\tabcolsep}{3.5pt}
    \begin{tabular}{lccc|ccc}
        \toprule
        Method & $\mathrm{AUC}_{20}$ & $T_{+}$ (k) & $T_{\mathrm{eval}+}$ (k) & $p_{\mathrm{AUC}_{20}}$ & $p_{T_{+}}$ & $p_{T_{\mathrm{eval}+}}$ \\
        \midrule
        AMSC & \(0.86\pm0.02\) & \(2.34\pm0.91\) & \(8.05\pm1.19\) & --- & --- & --- \\
        MOSAIC & \(0.35\pm0.03\) & \(5.83\pm0.84\) & \(14.44\pm1.35\) & 0.016 & 0.016 & 0.016 \\
        SER-s & \(0.91\pm0.04\) & \(1.59\pm0.72\) & \(7.12\pm1.38\) & 0.125 & 0.312 & 0.422 \\
        SER-r & \(0.86\pm0.05\) & \(3.27\pm1.98\) & \(10.10\pm2.65\) & 1.000 & 0.516 & 0.281 \\
        \bottomrule
    \end{tabular}
\end{table}

\begin{table}[h]
    \caption{Per-task CT28 early acquisition across seven matched seeds. The first four numeric columns are $\mathrm{AUC}_{20}$; the remaining four give $T_{+}$ in thousands of environment steps. Task $A$--D2 ($\dagger$) has no previous source and is excluded from the aggregate comparison.}
    \label{tbl:ct28_early_acquisition}
    \begin{center}
    \scriptsize
    \setlength{\tabcolsep}{3.3pt}
    \begin{tabular}{lcccc|cccc}
        \toprule
        & \multicolumn{4}{c}{$\mathrm{AUC}_{20}$} & \multicolumn{4}{c}{$T_{+}$ (k steps)} \\
        \cmidrule(lr){2-5}\cmidrule(lr){6-9}
        Task & AMSC & MOSAIC & SER-s & SER-r & AMSC & MOSAIC & SER-s & SER-r \\
        \midrule
        A--D2$^\dagger$ & 1.00 & 1.00 & 1.00 & 0.93 & 0.95 & 0.95 & 0.66 & 0.73 \\
        B--D2 & 0.79 & 1.00 & 0.64 & 0.79 & 0.66 & 0.73 & 2.56 & 1.76 \\
        C--D2 & 0.93 & 1.00 & 1.00 & 1.00 & 0.80 & 0.66 & 0.51 & 1.02 \\
        D--D2 & 0.79 & 0.86 & 0.71 & 0.86 & 1.10 & 1.46 & 1.68 & 1.17 \\
        \addlinespace
        A--D3 & 1.00 & 0.64 & 1.00 & 1.00 & 0.80 & 2.56 & 0.51 & 0.51 \\
        B--D3 & 1.00 & 0.29 & 1.00 & 1.00 & 0.95 & 1.39 & 0.51 & 0.51 \\
        C--D3 & 1.00 & 0.50 & 0.93 & 0.93 & 0.66 & 3.00 & 0.59 & 0.51 \\
        D--D3 & 0.93 & 1.00 & 1.00 & 1.00 & 0.88 & 1.46 & 0.51 & 0.51 \\
        \addlinespace
        A--D4 & 1.00 & 0.43 & 1.00 & 1.00 & 0.95 & 3.51 & 0.51 & 0.51 \\
        B--D4 & 1.00 & 0.50 & 1.00 & 1.00 & 0.80 & 4.17 & 0.51 & 0.51 \\
        C--D4 & 0.71 & 0.07 & 1.00 & 1.00 & 0.95 & 3.07 & 0.51 & 0.51 \\
        D--D4 & 1.00 & 0.36 & 0.86 & 1.00 & 1.10 & 4.90 & 0.51 & 0.51 \\
        \addlinespace
        A--D5 & 0.93 & 0.00 & 1.00 & 1.00 & 1.10 & 2.19 & 0.51 & 0.51 \\
        B--D5 & 1.00 & 0.21 & 1.00 & 1.00 & 1.02 & 6.88 & 0.51 & 0.51 \\
        C--D5 & 1.00 & 0.14 & 1.00 & 1.00 & 1.02 & 4.75 & 0.51 & 0.51 \\
        D--D5 & 1.00 & 0.14 & 0.93 & 1.00 & 0.80 & 5.19 & 1.24 & 0.51 \\
        \addlinespace
        A--D6 & 1.00 & 0.07 & 1.00 & 1.00 & 0.88 & 7.68 & 0.51 & 0.59 \\
        B--D6 & 1.00 & 0.14 & 1.00 & 1.00 & 1.02 & 7.17 & 0.51 & 0.51 \\
        C--D6 & 0.14 & 0.00 & 1.00 & 1.00 & 6.07 & 6.80 & 0.80 & 0.66 \\
        D--D6 & 1.00 & 0.43 & 0.57 & 0.93 & 0.88 & 6.44 & 5.92 & 0.51 \\
        \addlinespace
        A--D7 & 1.00 & 0.64 & 1.00 & 1.00 & 1.10 & 5.19 & 0.66 & 0.51 \\
        B--D7 & 0.93 & 0.00 & 0.79 & 0.29 & 1.02 & 8.27 & 0.80 & 8.70 \\
        C--D7 & 0.50 & 0.00 & 0.86 & 0.29 & 11.34 & 10.83 & 2.12 & 5.63 \\
        D--D7 & 1.00 & 0.43 & 0.79 & 0.86 & 1.02 & 10.02 & 7.97 & 8.05 \\
        \addlinespace
        A--D8 & 1.00 & 0.50 & 1.00 & 0.64 & 0.88 & 7.83 & 0.51 & 0.66 \\
        B--D8 & 0.71 & 0.00 & 0.93 & 0.43 & 1.24 & 10.90 & 0.66 & 15.07 \\
        C--D8 & 0.50 & 0.00 & 0.93 & 0.43 & 17.85 & 17.04 & 0.73 & 22.24 \\
        D--D8 & 0.43 & 0.07 & 0.64 & 0.79 & 6.29 & 13.39 & 10.02 & 15.07 \\
        \bottomrule
    \end{tabular}
    \end{center}
\end{table}


\subsection{Similarity-derived composition vs. learned composition in CW10}
\label{sec:cw10_analysis_extended}

\begin{figure}[h]
    \centering
    \includegraphics[width=\textwidth]
        {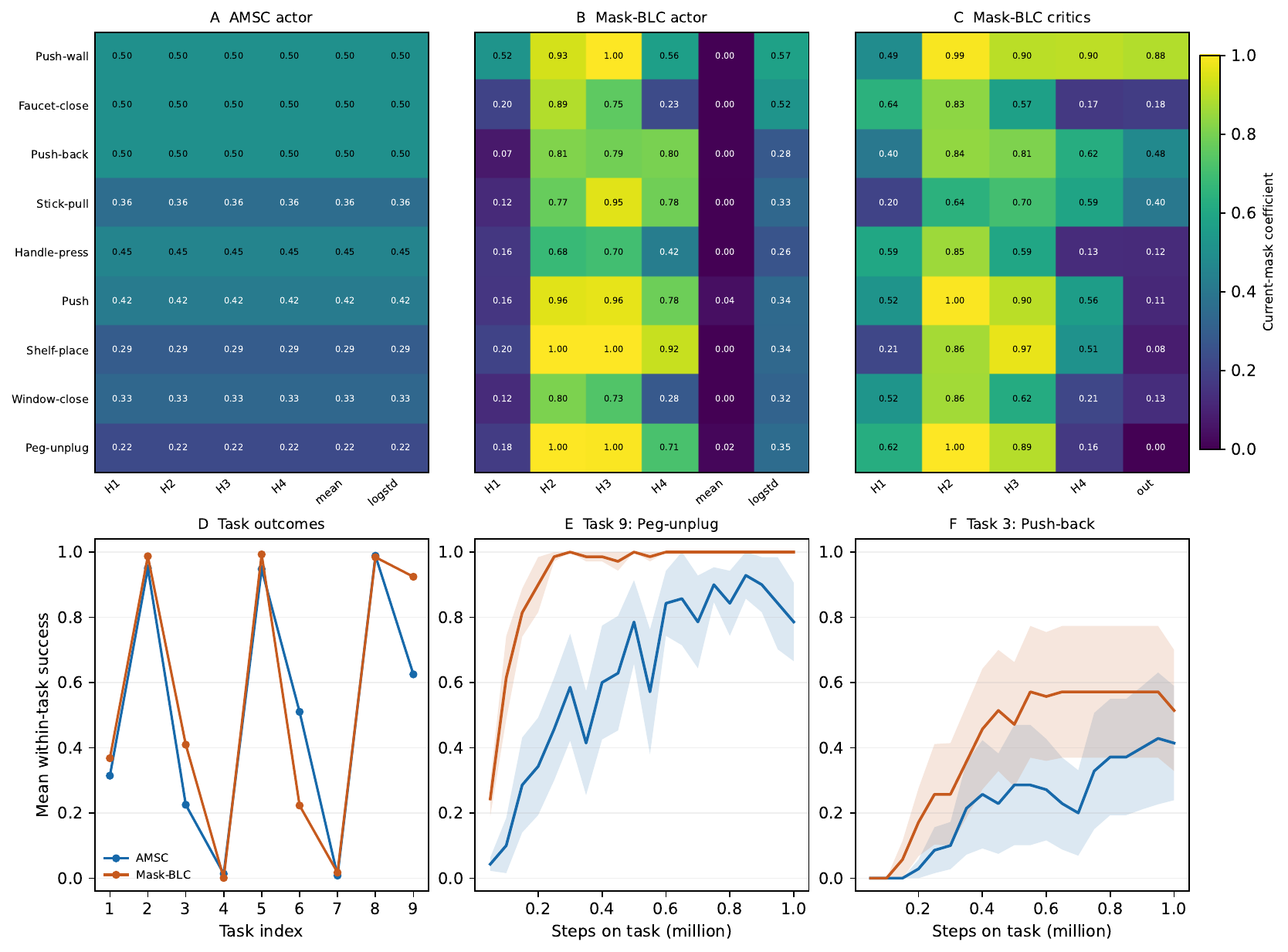}
    \caption{CW10 task-end composition and acquisition across seven seeds.
    Top: mean current-task mask coefficients by task and layer. Bottom:
    mean evaluation success by task and learning curves for the two tasks
    with the largest Mask-BLC--AMSC success gaps. Curves use 20 evaluation
    points per task. Coefficients describe the saved training-time
    allocation, not the effective contribution of each mask after
    task-boundary consolidation.}
    \label{fig:cw10_composition_layers}
\end{figure}

We compare the composition weights at the end of training of AMSC and
Mask-BLC. Figure~\ref{fig:cw10_composition_layers}
shows the coefficient assigned to the current-task mask at each layer,
alongside within-task evaluation success. AMSC uses one similarity-derived
allocation across its masked actor layers. Mask-BLC learns separate
weights for each actor and critic layer. Averaged over tasks 1--9,
Mask-BLC assigns approximately \(0.87\) to the current mask in its middle
actor layers, but only \(0.01\) in the action-mean head. Thus, a single
task-level similarity-derived weighting cannot recover the layer-dependent composition
learned by Mask-BLC in CW10.
\begin{figure}[h]
    \centering
    \includegraphics[width=0.9\textwidth]
        {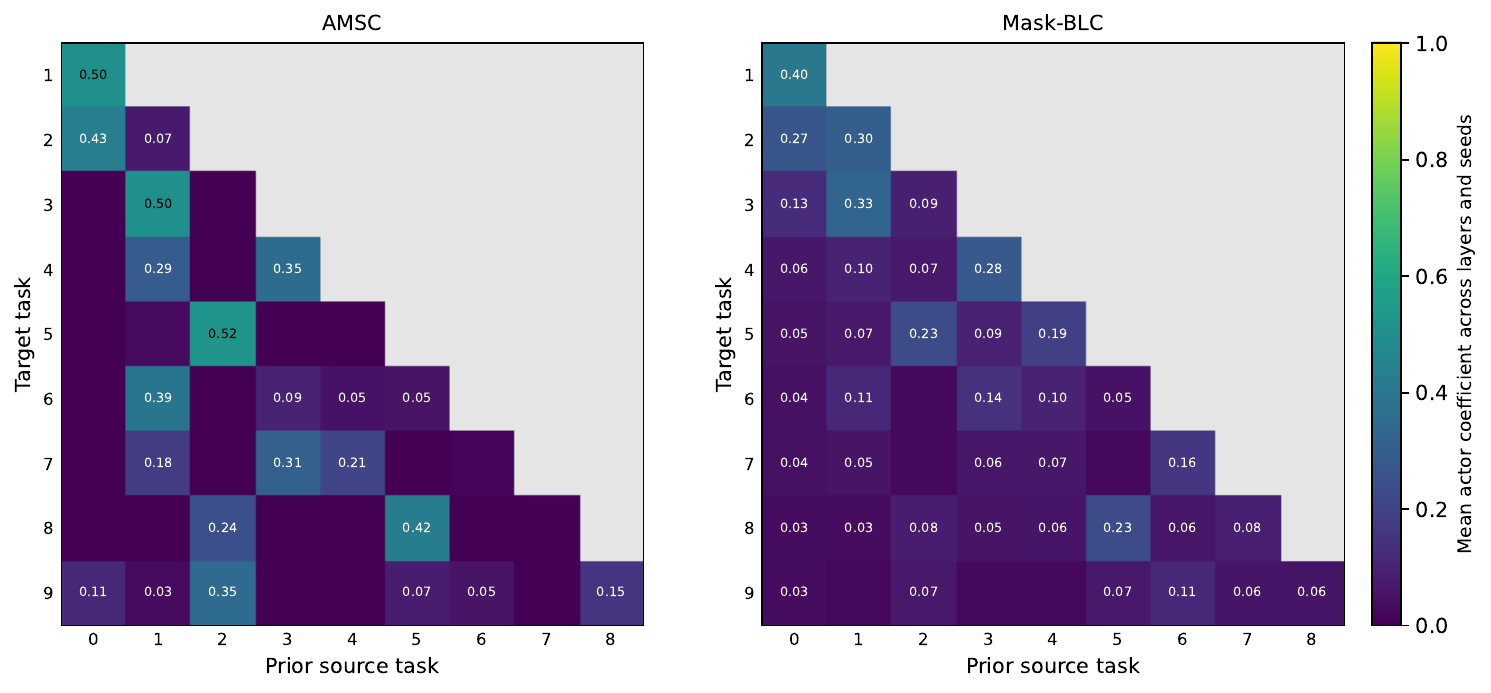}
    \caption{Mean actor coefficient assigned to each prior source task,
    averaged over layers and seven seeds. Rows are target tasks and
    columns are earlier source tasks; gray cells denote sources not yet
    available. Only coefficients of at least \(0.03\) are annotated.}
    \label{fig:cw10_source_allocation}
\end{figure}

The largest mean Mask-BLC--AMSC gaps occur on task 9
(peg-unplug-side, \(0.92\) versus \(0.63\) mean evaluation success) and task 3
(push-back, \(0.41\) versus \(0.23\)). The task-9 gap occurs in all seven
paired seeds, whereas the task-3 gap is less consistent. AMSC also exceeds
Mask-BLC on task 6 (\(0.51\) versus \(0.22\)), so the difference is not
uniform across the curriculum. Task 3 remains above its from-scratch
reference; it explains part of the between-method performance gap, but
not AMSC's negative forward transfer. Normalized FWT can additionally
amplify small deficits on tasks whose from-scratch performance is close
to one.

Figure~\ref{fig:cw10_source_allocation} identifies the prior tasks receiving
actor coefficient weight. AMSC's sparse support excludes some sources
entirely, while Mask-BLC distributes learned coefficients over the
available library. 

\subsection{Hyperparameter Sensitivity}
\label{sec:sensitivity_analysis}
Each sweep changes one AMSC hyperparameter while retaining a
benchmark-specific sweep reference for all others. Table~\ref{tbl:amsc_sensitivity_ranges} reports every value considered. Frequencies are measured in PPO iterations, and bold values denote the sweep-reference settings. Each setting contains seven seeds. The reported performance intervals are the minimum and maximum of the seven-seed means across the swept values, rather than confidence intervals.
\begin{table*}[h]
    \caption{Sensitivity ranges for AMSC. Each AUC/FWT interval is
    the range of mean performance across the values in the adjacent sweep.
    All settings obtain zero measured FGT. \(N\) is the number of task samples,
    \(R\) is the reference-distribution size, and \(u\) is the reference seed.
    MG16 uses \(N=512\) outside the sample-count sweep. Bold marks sweep
    references.}
    \label{tbl:amsc_sensitivity_ranges}
    \begin{center}
    \scriptsize
    \setlength{\tabcolsep}{3.5pt}
    \begin{tabularx}{\textwidth}{lXccXcc}
        \toprule
        & \multicolumn{3}{c}{\textbf{CT28}} &
          \multicolumn{3}{c}{\textbf{MG16}} \\
        \cmidrule(lr){2-4}\cmidrule(lr){5-7}
        \textbf{Parameter} & \textbf{Swept values} &
        \textbf{AUC} & \textbf{FWT} & \textbf{Swept values} &
        \textbf{AUC} & \textbf{FWT} \\
        \midrule
        Temperature \(\tau\)
        & \(0.25,0.5,\mathbf{1},2,4\) & \(0.89\text{--}0.96\) & \(0.88\text{--}0.95\)
        & \(0.25,0.5,\mathbf{1},2,4\) & \(0.50\text{--}0.67\) & \(0.16\text{--}0.50\) \\
        Samples \(N\)
        & \(32,64,\mathbf{128},256,512\) & \(0.94\text{--}0.95\) & \(0.93\text{--}0.95\)
        & \(128,256,512,\allowbreak\mathbf{1024},2048\) & \(0.54\text{--}0.73\) & \(0.23\text{--}0.54\) \\
        Reference size \(R\)
        & \(10,25,\mathbf{50},100,200\) & \(0.89\text{--}0.94\) & \(0.88\text{--}0.94\)
        & \(10,25,\mathbf{50},100,200\) & \(0.55\text{--}0.70\) & \(0.35\text{--}0.52\) \\
        Embedding period \(f_{\mathrm{emb}}\)
        & \(\mathbf{1},2,5,10\) & \(0.84\text{--}0.94\) & \(0.83\text{--}0.94\)
        & \(\mathbf{1},2,5,10\) & \(0.57\text{--}0.62\) & \(0.34\text{--}0.43\) \\
        Selection period \(f_{\mathrm{sel}}\)
        & \(\mathbf{1},2,5,10\) & \(0.83\text{--}0.94\) & \(0.81\text{--}0.94\)
        & \(1,2,\mathbf{5},10\) & \(0.57\text{--}0.65\) & \(0.35\text{--}0.43\) \\
        EMA coefficient \(\rho\)
        & \(0,0.25,\mathbf{0.5},0.75,0.9\) & \(0.93\text{--}0.95\) & \(0.92\text{--}0.95\)
        & \(0,0.25,\mathbf{0.5},0.75,0.9\) & \(0.62\text{--}0.68\) & \(0.43\text{--}0.51\) \\
        Reference seed \(u\)
        & \(11,37,71,\mathbf{98},131\) & \(0.92\text{--}0.96\) & \(0.91\text{--}0.96\)
        & \(11,37,71,\mathbf{98},131\) & \(0.62\text{--}0.68\) & \(0.39\text{--}0.50\) \\
        \bottomrule
    \end{tabularx}
    \end{center}
\end{table*}

Tables~\ref{tbl:amsc_sensitivity_temperature}--
\ref{tbl:amsc_sensitivity_reference_seed} report the complete setting-level
results, including bootstrap confidence intervals.

\begin{table}[h]
    \caption{Sparsemax-temperature sensitivity. Values are means with asymmetric 95\% bootstrap confidence intervals across seven seeds. Bold entries denote the sweep-reference setting; All runs obtain zero measured FGT.}
    \label{tbl:amsc_sensitivity_temperature}
    \begin{center}
    \scriptsize
    \setlength{\tabcolsep}{4pt}
    \begin{tabular}{lccc|lccc}
        \toprule
        \multicolumn{4}{c}{\textbf{CT28}} & \multicolumn{4}{c}{\textbf{MG16}} \\
        \cmidrule(lr){1-4}\cmidrule(lr){5-8}
        $\tau$ & AUC & FWT & FGT & $\tau$ & AUC & FWT & FGT \\
        \midrule
        0.25 & $0.93_{-0.04}^{+0.04}$ & $0.92_{-0.05}^{+0.05}$ & $0.00$ & 0.25 & $0.63_{-0.08}^{+0.07}$ & $0.45_{-0.08}^{+0.09}$ & $0.00$ \\
        0.5 & $0.96_{-0.03}^{+0.03}$ & $0.95_{-0.04}^{+0.04}$ & $0.00$ & 0.5 & $0.67_{-0.09}^{+0.10}$ & $0.50_{-0.11}^{+0.12}$ & $0.00$ \\
        $\mathbf{1}$ & $\mathbf{0.94_{-0.02}^{+0.02}}$ & $\mathbf{0.94_{-0.03}^{+0.02}}$ & $\mathbf{0.00}$ & $\mathbf{1}$ & $\mathbf{0.62_{-0.09}^{+0.08}}$ & $\mathbf{0.43_{-0.12}^{+0.13}}$ & $\mathbf{0.00}$ \\
        2 & $0.93_{-0.03}^{+0.03}$ & $0.92_{-0.04}^{+0.03}$ & $0.00$ & 2 & $0.54_{-0.12}^{+0.10}$ & $0.31_{-0.14}^{+0.12}$ & $0.00$ \\
        4 & $0.89_{-0.02}^{+0.03}$ & $0.88_{-0.03}^{+0.04}$ & $0.00$ & 4 & $0.50_{-0.07}^{+0.07}$ & $0.16_{-0.22}^{+0.18}$ & $0.00$ \\
        \bottomrule
    \end{tabular}
    \end{center}
\end{table}
\begin{table}[h]
    \caption{Task-sample-count sensitivity. Values are means with asymmetric 95\% bootstrap confidence intervals across seven seeds. Bold entries denote the principal sample counts (CT28: $N=128$; MG16: $N=1024$). All runs obtain zero measured FGT.}
    \label{tbl:amsc_sensitivity_sample_count}
    \begin{center}
    \scriptsize
    \setlength{\tabcolsep}{4pt}
    \begin{tabular}{lccc|lccc}
        \toprule
        \multicolumn{4}{c}{\textbf{CT28}} & \multicolumn{4}{c}{\textbf{MG16}} \\
        \cmidrule(lr){1-4}\cmidrule(lr){5-8}
        $N$ & AUC & FWT & FGT & $N$ & AUC & FWT & FGT \\
        \midrule
        32 & $0.94_{-0.04}^{+0.03}$ & $0.93_{-0.04}^{+0.04}$ & $0.00$ & 128 & $0.54_{-0.12}^{+0.11}$ & $0.23_{-0.28}^{+0.22}$ & $0.00$ \\
        64 & $0.95_{-0.03}^{+0.02}$ & $0.95_{-0.03}^{+0.03}$ & $0.00$ & 256 & $0.57_{-0.13}^{+0.12}$ & $0.33_{-0.18}^{+0.17}$ & $0.00$ \\
        $\mathbf{128}$ & $\mathbf{0.94_{-0.02}^{+0.02}}$ & $\mathbf{0.94_{-0.03}^{+0.02}}$ & $\mathbf{0.00}$ & $512$ & $0.62_{-0.09}^{+0.08}$ & $0.43_{-0.12}^{+0.13}$ & $0.00$ \\
        256 & $0.95_{-0.03}^{+0.03}$ & $0.95_{-0.04}^{+0.03}$ & $0.00$ & $\mathbf{1024}$ & $\mathbf{0.73_{-0.06}^{+0.06}}$ & $\mathbf{0.54_{-0.14}^{+0.12}}$ & $\mathbf{0.00}$ \\
        512 & $0.95_{-0.04}^{+0.03}$ & $0.95_{-0.03}^{+0.03}$ & $0.00$ & 2048 & $0.63_{-0.13}^{+0.10}$ & $0.30_{-0.44}^{+0.29}$ & $0.00$ \\
        \bottomrule
    \end{tabular}
    \end{center}
\end{table}
\begin{table}[h]
    \caption{Reference-distribution-size sensitivity. Values are means with asymmetric 95\% bootstrap confidence intervals across seven seeds. Bold entries denote the sweep-reference setting. All runs obtain zero measured FGT.}
    \label{tbl:amsc_sensitivity_reference_size}
    \begin{center}
    \scriptsize
    \setlength{\tabcolsep}{4pt}
    \begin{tabular}{lccc|lccc}
        \toprule
        \multicolumn{4}{c}{\textbf{CT28}} & \multicolumn{4}{c}{\textbf{MG16}} \\
        \cmidrule(lr){1-4}\cmidrule(lr){5-8}
        $R$ & AUC & FWT & FGT & $R$ & AUC & FWT & FGT \\
        \midrule
        10 & $0.94_{-0.04}^{+0.03}$ & $0.94_{-0.05}^{+0.04}$ & $0.00$ & 10 & $0.70_{-0.09}^{+0.08}$ & $0.52_{-0.14}^{+0.14}$ & $0.00$ \\
        25 & $0.94_{-0.03}^{+0.03}$ & $0.93_{-0.04}^{+0.03}$ & $0.00$ & 25 & $0.55_{-0.09}^{+0.09}$ & $0.35_{-0.16}^{+0.16}$ & $0.00$ \\
        $\mathbf{50}$ & $\mathbf{0.94_{-0.02}^{+0.02}}$ & $\mathbf{0.94_{-0.03}^{+0.02}}$ & $\mathbf{0.00}$ & $\mathbf{50}$ & $\mathbf{0.62_{-0.09}^{+0.08}}$ & $\mathbf{0.43_{-0.12}^{+0.13}}$ & $\mathbf{0.00}$ \\
        100 & $0.89_{-0.02}^{+0.03}$ & $0.88_{-0.03}^{+0.04}$ & $0.00$ & 100 & $0.69_{-0.10}^{+0.10}$ & $0.50_{-0.12}^{+0.14}$ & $0.00$ \\
        200 & $0.91_{-0.03}^{+0.03}$ & $0.90_{-0.03}^{+0.03}$ & $0.00$ & 200 & $0.66_{-0.11}^{+0.10}$ & $0.48_{-0.14}^{+0.14}$ & $0.00$ \\
        \bottomrule
    \end{tabular}
    \end{center}
\end{table}
\begin{table}[h]
    \caption{Embedding-update-period sensitivity. Values are means with asymmetric 95\% bootstrap confidence intervals across seven seeds. Bold entries denote the sweep-reference setting. All runs obtain zero measured FGT.}
    \label{tbl:amsc_sensitivity_embedding_frequency}
    \begin{center}
    \scriptsize
    \setlength{\tabcolsep}{4pt}
    \begin{tabular}{lccc|lccc}
        \toprule
        \multicolumn{4}{c}{\textbf{CT28}} & \multicolumn{4}{c}{\textbf{MG16}} \\
        \cmidrule(lr){1-4}\cmidrule(lr){5-8}
        $f_{\mathrm{emb}}$ & AUC & FWT & FGT & $f_{\mathrm{emb}}$ & AUC & FWT & FGT \\
        \midrule
        $\mathbf{1}$ & $\mathbf{0.94_{-0.02}^{+0.02}}$ & $\mathbf{0.94_{-0.03}^{+0.02}}$ & $\mathbf{0.00}$ & $\mathbf{1}$ & $\mathbf{0.62_{-0.09}^{+0.08}}$ & $\mathbf{0.43_{-0.12}^{+0.13}}$ & $\mathbf{0.00}$ \\
        2 & $0.91_{-0.03}^{+0.03}$ & $0.91_{-0.03}^{+0.04}$ & $0.00$ & 2 & $0.59_{-0.15}^{+0.14}$ & $0.41_{-0.19}^{+0.18}$ & $0.00$ \\
        5 & $0.90_{-0.03}^{+0.03}$ & $0.90_{-0.05}^{+0.04}$ & $0.00$ & 5 & $0.58_{-0.15}^{+0.13}$ & $0.36_{-0.25}^{+0.21}$ & $0.00$ \\
        10 & $0.84_{-0.03}^{+0.03}$ & $0.83_{-0.05}^{+0.05}$ & $0.00$ & 10 & $0.57_{-0.13}^{+0.13}$ & $0.34_{-0.13}^{+0.14}$ & $0.00$ \\
        \bottomrule
    \end{tabular}
    \end{center}
\end{table}
\begin{table}[h]
    \caption{Selection-update-period sensitivity. Values are means with asymmetric 95\% bootstrap confidence intervals across seven seeds. Bold entries denote the sweep-reference setting; MG16 uses $N=512$ outside the sample-count sweep. All runs obtain zero measured FGT.}
    \label{tbl:amsc_sensitivity_selection_frequency}
    \begin{center}
    \scriptsize
    \setlength{\tabcolsep}{4pt}
    \begin{tabular}{lccc|lccc}
        \toprule
        \multicolumn{4}{c}{\textbf{CT28}} & \multicolumn{4}{c}{\textbf{MG16}} \\
        \cmidrule(lr){1-4}\cmidrule(lr){5-8}
        $f_{\mathrm{sel}}$ & AUC & FWT & FGT & $f_{\mathrm{sel}}$ & AUC & FWT & FGT \\
        \midrule
        $\mathbf{1}$ & $\mathbf{0.94_{-0.02}^{+0.02}}$ & $\mathbf{0.94_{-0.03}^{+0.02}}$ & $\mathbf{0.00}$ & 1 & $0.64_{-0.08}^{+0.08}$ & $0.40_{-0.20}^{+0.18}$ & $0.00$ \\
        2 & $0.91_{-0.02}^{+0.03}$ & $0.90_{-0.03}^{+0.03}$ & $0.00$ & 2 & $0.57_{-0.15}^{+0.14}$ & $0.35_{-0.16}^{+0.14}$ & $0.00$ \\
        5 & $0.86_{-0.02}^{+0.02}$ & $0.85_{-0.02}^{+0.02}$ & $0.00$ & $\mathbf{5}$ & $\mathbf{0.62_{-0.09}^{+0.08}}$ & $\mathbf{0.43_{-0.12}^{+0.13}}$ & $\mathbf{0.00}$ \\
        10 & $0.83_{-0.02}^{+0.03}$ & $0.81_{-0.04}^{+0.04}$ & $0.00$ & 10 & $0.65_{-0.09}^{+0.08}$ & $0.42_{-0.13}^{+0.12}$ & $0.00$ \\
        \bottomrule
    \end{tabular}
    \end{center}
\end{table}
\begin{table}[h]
    \caption{Embedding-EMA sensitivity. Values are means with asymmetric 95\% bootstrap confidence intervals across seven seeds. Bold entries denote the sweep-reference setting; MG16 uses $N=512$ outside the sample-count sweep. All runs obtain zero measured FGT.}
    \label{tbl:amsc_sensitivity_ema}
    \begin{center}
    \scriptsize
    \setlength{\tabcolsep}{4pt}
    \begin{tabular}{lccc|lccc}
        \toprule
        \multicolumn{4}{c}{\textbf{CT28}} & \multicolumn{4}{c}{\textbf{MG16}} \\
        \cmidrule(lr){1-4}\cmidrule(lr){5-8}
        $\rho$ & AUC & FWT & FGT & $\rho$ & AUC & FWT & FGT \\
        \midrule
        0 & $0.95_{-0.03}^{+0.03}$ & $0.94_{-0.04}^{+0.03}$ & $0.00$ & 0 & $0.68_{-0.12}^{+0.10}$ & $0.51_{-0.14}^{+0.14}$ & $0.00$ \\
        0.25 & $0.95_{-0.03}^{+0.02}$ & $0.95_{-0.03}^{+0.02}$ & $0.00$ & 0.25 & $0.66_{-0.10}^{+0.10}$ & $0.48_{-0.14}^{+0.14}$ & $0.00$ \\
        $\mathbf{0.5}$ & $\mathbf{0.94_{-0.02}^{+0.02}}$ & $\mathbf{0.94_{-0.03}^{+0.02}}$ & $\mathbf{0.00}$ & $\mathbf{0.5}$ & $\mathbf{0.62_{-0.09}^{+0.08}}$ & $\mathbf{0.43_{-0.12}^{+0.13}}$ & $\mathbf{0.00}$ \\
        0.75 & $0.93_{-0.03}^{+0.03}$ & $0.92_{-0.03}^{+0.03}$ & $0.00$ & 0.75 & $0.67_{-0.08}^{+0.09}$ & $0.48_{-0.10}^{+0.12}$ & $0.00$ \\
        0.9 & $0.93_{-0.03}^{+0.03}$ & $0.92_{-0.04}^{+0.04}$ & $0.00$ & 0.9 & $0.66_{-0.13}^{+0.12}$ & $0.47_{-0.17}^{+0.17}$ & $0.00$ \\
        \bottomrule
    \end{tabular}
    \end{center}
\end{table}
\begin{table}[h]
    \caption{Random-reference-seed sensitivity. Values are means with asymmetric 95\% bootstrap confidence intervals across seven seeds. Bold entries denote the sweep-reference setting; MG16 uses $N=512$ outside the sample-count sweep. All runs obtain zero measured FGT.}
    \label{tbl:amsc_sensitivity_reference_seed}
    \begin{center}
    \scriptsize
    \setlength{\tabcolsep}{4pt}
    \begin{tabular}{lccc|lccc}
        \toprule
        \multicolumn{4}{c}{\textbf{CT28}} & \multicolumn{4}{c}{\textbf{MG16}} \\
        \cmidrule(lr){1-4}\cmidrule(lr){5-8}
        $u$ & AUC & FWT & FGT & $u$ & AUC & FWT & FGT \\
        \midrule
        11 & $0.96_{-0.03}^{+0.02}$ & $0.96_{-0.03}^{+0.02}$ & $0.00$ & 11 & $0.65_{-0.08}^{+0.08}$ & $0.47_{-0.13}^{+0.12}$ & $0.00$ \\
        37 & $0.93_{-0.03}^{+0.03}$ & $0.92_{-0.04}^{+0.04}$ & $0.00$ & 37 & $0.65_{-0.09}^{+0.09}$ & $0.46_{-0.15}^{+0.14}$ & $0.00$ \\
        71 & $0.92_{-0.03}^{+0.03}$ & $0.91_{-0.03}^{+0.04}$ & $0.00$ & 71 & $0.62_{-0.09}^{+0.08}$ & $0.39_{-0.14}^{+0.14}$ & $0.00$ \\
        $\mathbf{98}$ & $\mathbf{0.94_{-0.02}^{+0.02}}$ & $\mathbf{0.94_{-0.03}^{+0.02}}$ & $\mathbf{0.00}$ & $\mathbf{98}$ & $\mathbf{0.62_{-0.09}^{+0.08}}$ & $\mathbf{0.43_{-0.12}^{+0.13}}$ & $\mathbf{0.00}$ \\
        131 & $0.92_{-0.03}^{+0.03}$ & $0.91_{-0.04}^{+0.04}$ & $0.00$ & 131 & $0.68_{-0.12}^{+0.10}$ & $0.50_{-0.17}^{+0.16}$ & $0.00$ \\
        \bottomrule
    \end{tabular}
    \end{center}
\end{table}

\paragraph{Representation construction and retrieval.}
The CT28 reference records AUC/FWT of \(0.94/0.94\) and remains stable, with both metrics remaining between \(0.93\) and \(0.95\). Moderate temperature changes are also tolerated. \(\tau=0.5\) records \(0.96/0.95\), whereas flattening the retrieval distribution with \(\tau=4\) reduces the result to \(0.89/0.88\). Reference sizes of 10, 25, and 50 remain near \(0.94/0.94\), while sizes of 100 and 200 obtain \(0.89/0.88\) and \(0.91/0.90\), respectively.

MG16 is more sensitive to representation quality. The independently repeated
reference configuration records \(0.62/0.43\); its difference from the
earlier \(N=512\) reference run (\(0.62/0.37\)) is not significant
(two-sided paired \(p=0.92/0.41\)). Lowering \(\tau\) to 0.5 raises the sweep
mean to \(0.67/0.50\), while \(\tau=4\) obtains \(0.50/0.16\). Increasing
\(N\) from 128 to 1024 changes AUC/FWT from \(0.54/0.23\) to \(0.73/0.54\),
but 2048 samples reduce FWT to \(0.30\). Reference-size results are likewise
non-monotonic: \(R=10\) obtains \(0.70/0.52\), while \(R=25\) obtains
\(0.55/0.35\). These point estimates show that more samples do not
monotonically improve acquisition.

\paragraph{Update schedule and smoothing.}
Frequent updates are most important on CT28. Increasing either
\(f_{\mathrm{emb}}\) or \(f_{\mathrm{sel}}\) from 1 to 10 reduces AUC/FWT
from \(0.94/0.94\) to \(0.84/0.83\) and \(0.83/0.81\), respectively. On
MG16, increasing \(f_{\mathrm{emb}}\) from 1 to 10 changes \(0.62/0.43\) to
\(0.57/0.34\). Selection frequency is less ordered: the sweep-reference period of 5
has the highest FWT (0.43), whereas a period of 10 has the highest AUC (0.65).
EMA smoothing is not required for the best point estimate in either sweep.
CT28 peaks at \(\rho=0.25\) with \(0.95/0.95\), and MG16 peaks without EMA at
\(0.68/0.51\). Changing the random reference seed produces AUC ranges of
\(0.92\text{--}0.96\) and \(0.62\text{--}0.68\), and FWT ranges of
\(0.91\text{--}0.96\) and \(0.39\text{--}0.50\), on CT28 and MG16,
respectively. These sweeps describe one-factor sensitivity around the stated references;
they do not establish robustness to joint hyperparameter changes or document
how the principal configuration was selected. Retention is unaffected across all settings.
The observed sensitivity concerns acquisition.

\subsection{Per-Task Acquisition and Retention}
\label{sec:per_task_metrics}
Tables~\ref{tbl:per_task_ct28_amsc_masklc} and
\ref{tbl:per_task_mg16_amsc_masklc} report task-level AUC, FWT, and FGT for
AMSC and full-library learned composition.
Table~\ref{tbl:per_task_cw10_amsc_masklc_maskblc} additionally reports CW10
comparisons with Mask-LC and Mask-BLC. These results expose variation
hidden by curriculum averages. The supplied MG16 per-task table summarizes
an earlier result set: averaging its displayed task AUCs gives approximately
\(0.62\), rather than the principal \(0.73\). It is retained as a separately
identified analysis and must not be used as a decomposition of the current
main-text aggregate. CT28 gains are broad but diminish on several
late, high-depth tasks. On MG16, both methods exhibit negative normalized FWT
on some R2 cold-start tasks, while AMSC improves acquisition across the later
R3--R5 tasks. Both parameter-isolation methods retain every completed task,
yielding zero measured FGT throughout.


\begin{table*}[h]
    \caption{Per-task CT28 acquisition and retention. Values are means across seven seeds. FWT is undefined when the single-task expert saturates the normalization denominator and can fall below $-1$ when the expert's residual headroom is small.}
    \label{tbl:per_task_ct28_amsc_masklc}
    \begin{center}
    \scriptsize
    \setlength{\tabcolsep}{4pt}
    \begin{tabular}{lccc|ccc}
        \toprule
        & \multicolumn{3}{c}{\textbf{AMSC}} & \multicolumn{3}{c}{\textbf{Mask-LC}} \\
        \cmidrule(lr){2-4}\cmidrule(lr){5-7}
        Task & AUC & FWT & FGT & AUC & FWT & FGT \\
        \midrule
        A--D2 & 1.00 & --- & 0.00 & 1.00 & --- & 0.00 \\
        B--D2 & 0.96 & --- & 0.00 & 0.97 & --- & 0.00 \\
        C--D2 & 0.99 & --- & 0.00 & 1.00 & --- & 0.00 \\
        D--D2 & 0.96 & 0.00 & 0.00 & 0.89 & 0.00 & 0.00 \\
        A--D3 & 1.00 & 1.00 & 0.00 & 1.00 & 1.00 & 0.00 \\
        B--D3 & 1.00 & 1.00 & 0.00 & 0.80 & 0.63 & 0.00 \\
        C--D3 & 1.00 & 1.00 & 0.00 & 0.99 & 0.95 & 0.00 \\
        D--D3 & 0.99 & 0.98 & 0.00 & 0.59 & -0.28 & 0.00 \\
        A--D4 & 1.00 & 1.00 & 0.00 & 0.97 & 0.98 & 0.00 \\
        B--D4 & 1.00 & 1.00 & 0.00 & 0.66 & 0.67 & 0.00 \\
        C--D4 & 0.94 & 0.97 & 0.00 & 0.39 & 0.40 & 0.00 \\
        D--D4 & 1.00 & 1.00 & 0.00 & 0.37 & 0.37 & 0.00 \\
        A--D5 & 0.99 & 0.99 & 0.00 & 0.76 & 0.79 & 0.00 \\
        B--D5 & 1.00 & 1.00 & 0.00 & 0.34 & 0.34 & 0.00 \\
        C--D5 & 1.00 & 1.00 & 0.00 & 0.41 & 0.42 & 0.00 \\
        D--D5 & 1.00 & 1.00 & 0.00 & 0.00 & 0.00 & 0.00 \\
        A--D6 & 1.00 & 1.00 & 0.00 & 0.84 & 0.88 & 0.00 \\
        B--D6 & 1.00 & 1.00 & 0.00 & 0.00 & 0.00 & 0.00 \\
        C--D6 & 0.60 & 0.63 & 0.00 & 0.00 & 0.00 & 0.00 \\
        D--D6 & 1.00 & 1.00 & 0.00 & 0.00 & 0.00 & 0.00 \\
        A--D7 & 1.00 & 1.00 & 0.00 & 0.69 & 0.71 & 0.00 \\
        B--D7 & 0.99 & 0.99 & 0.00 & 0.00 & 0.00 & 0.00 \\
        C--D7 & 0.56 & 0.56 & 0.00 & 0.00 & 0.00 & 0.00 \\
        D--D7 & 1.00 & 1.00 & 0.00 & 0.00 & 0.00 & 0.00 \\
        A--D8 & 1.00 & 1.00 & 0.00 & 0.66 & 0.68 & 0.00 \\
        B--D8 & 0.94 & 0.97 & 0.00 & 0.00 & 0.00 & 0.00 \\
        C--D8 & 0.67 & 0.68 & 0.00 & 0.00 & 0.00 & 0.00 \\
        D--D8 & 0.77 & 0.81 & 0.00 & 0.00 & 0.00 & 0.00 \\
        \bottomrule
    \end{tabular}
    \end{center}
\end{table*}
\begin{table*}[h]
    \caption{Per-task MG16 acquisition and retention for the earlier result set (mean displayed AUC approximately $0.62$). This table does not decompose the principal $0.73$ AUC result. Values are means across seven seeds. FWT is undefined when the single-task expert saturates the normalization denominator and can fall below $-1$ when the expert's residual headroom is small.}
    \label{tbl:per_task_mg16_amsc_masklc}
    \begin{center}
    \scriptsize
    \setlength{\tabcolsep}{4pt}
    \begin{tabular}{lccc|ccc}
        \toprule
        & \multicolumn{3}{c}{\textbf{AMSC}} & \multicolumn{3}{c}{\textbf{Mask-LC}} \\
        \cmidrule(lr){2-4}\cmidrule(lr){5-7}
        Task & AUC & FWT & FGT & AUC & FWT & FGT \\
        \midrule
        BallLava--R2 & 0.83 & 0.24 & 0.00 & 0.81 & 0.23 & 0.00 \\
        LavaBall--R2 & 0.81 & -1.27 & 0.00 & 0.78 & -2.41 & 0.00 \\
        KeyBoxSwap--R2 & 0.91 & 0.21 & 0.00 & 0.81 & -2.14 & 0.00 \\
        Identity--R2 & 0.79 & 0.21 & 0.00 & 0.40 & -2.57 & 0.00 \\
        BallLava--R3 & 0.72 & 0.72 & 0.00 & 0.63 & 0.63 & 0.00 \\
        LavaBall--R3 & 0.56 & 0.56 & 0.00 & 0.44 & 0.44 & 0.00 \\
        KeyBoxSwap--R3 & 0.74 & 0.72 & 0.00 & 0.47 & 0.47 & 0.00 \\
        Identity--R3 & 0.50 & 0.46 & 0.00 & 0.13 & -0.42 & 0.00 \\
        BallLava--R4 & 0.62 & 0.62 & 0.00 & 0.36 & 0.36 & 0.00 \\
        LavaBall--R4 & 0.69 & 0.69 & 0.00 & 0.33 & 0.33 & 0.00 \\
        KeyBoxSwap--R4 & 0.45 & 0.45 & 0.00 & 0.42 & 0.42 & 0.00 \\
        Identity--R4 & 0.24 & 0.24 & 0.00 & 0.18 & 0.18 & 0.00 \\
        BallLava--R5 & 0.61 & 0.61 & 0.00 & 0.37 & 0.38 & 0.00 \\
        LavaBall--R5 & 0.57 & 0.57 & 0.00 & 0.13 & 0.13 & 0.00 \\
        KeyBoxSwap--R5 & 0.49 & 0.49 & 0.00 & 0.13 & 0.13 & 0.00 \\
        Identity--R5 & 0.38 & 0.39 & 0.00 & 0.06 & 0.06 & 0.00 \\
        \bottomrule
    \end{tabular}
    \end{center}
\end{table*}


\begin{table*}[h]
    \caption{Per-task CW10 acquisition and retention across seven shared seeds. AUC averages the 20 active-task evaluations; FWT compares trapezoidal areas with the seed-matched from-scratch run. FWT is undefined when the reference headroom is below $10^{-3}$ and can be strongly negative when that headroom is small. FGT compares task-end and final success. Window-close FWT uses only one non-saturated seed; its AUC and FGT and all other rows use seven seeds.}
    \label{tbl:per_task_cw10_amsc_masklc_maskblc}
    \begin{center}
    \scriptsize
    \setlength{\tabcolsep}{3pt}
    \begin{tabular}{lccc|ccc|ccc}
        \toprule
        & \multicolumn{3}{c}{\textbf{AMSC}} & \multicolumn{3}{c}{\textbf{Mask-LC}} & \multicolumn{3}{c}{\textbf{Mask-BLC}} \\
        \cmidrule(lr){2-4}\cmidrule(lr){5-7}\cmidrule(lr){8-10}
        Task & AUC & FWT & FGT & AUC & FWT & FGT & AUC & FWT & FGT \\
        \midrule
        0: Hammer & 0.24 & -1.62 & 0.00 & 0.36 & -0.67 & 0.00 & 0.36 & -0.67 & 0.00 \\
        1: Push-wall & 0.31 & -0.05 & -0.01 & 0.37 & 0.08 & 0.00 & 0.37 & 0.08 & 0.00 \\
        2: Faucet-close & 0.95 & -3.27 & 0.00 & 0.99 & 0.22 & 0.00 & 0.99 & 0.13 & 0.00 \\
        3: Push-back & 0.23 & 0.23 & 0.00 & 0.44 & 0.44 & -0.01 & 0.41 & 0.42 & 0.00 \\
        4: Stick-pull & 0.01 & -0.07 & 0.00 & 0.00 & -0.09 & 0.00 & 0.00 & -0.09 & 0.00 \\
        5: Handle-press-side & 0.95 & -4.67 & 0.00 & 0.99 & 0.93 & 0.00 & 0.99 & 0.43 & 0.00 \\
        6: Push & 0.51 & 0.43 & 0.00 & 0.09 & -0.11 & -0.01 & 0.22 & 0.06 & 0.00 \\
        7: Shelf-place & 0.01 & 0.01 & -0.01 & 0.00 & 0.00 & 0.00 & 0.02 & 0.02 & 0.00 \\
        8: Window-close & 0.99 & 1.00 & 0.00 & 0.98 & 0.80 & 0.00 & 0.98 & 0.50 & 0.00 \\
        9: Peg-unplug-side & 0.62 & -2.09 & 0.00 & 0.53 & -2.92 & 0.00 & 0.92 & 0.60 & 0.00 \\
        \bottomrule
    \end{tabular}
    \end{center}
\end{table*}

\section{Evaluation Metrics}
\label{sec:metrics}

Consider a curriculum of \(N\) tasks with an equal interaction budget
\(\Delta\) per task and total budget \(T=N\Delta\). Task \(i\) is trained
during \([t_i,e_i]\), where \(t_i=(i-1)\Delta\) and \(e_i=i\Delta\).
Let \(p_i(t)\in[0,1]\) be the normalized evaluation performance on task
\(i\) at checkpoint \(t\). Each checkpoint value averages the evaluation
episodes conducted without training.

CT28 returns and CW10 success rates already lie in \([0,1]\). For MG16, we
divide each return by the corresponding task's theoretical shortest-path
return and clip the result to \([0,1]\). The same task-specific normalization
is applied to lifelong and single-task runs. Evaluations made before a task is
first encountered are excluded from acquisition metrics.

\paragraph{Acquisition AUC.}
Suppose task \(i\)'s active training block contains \(M_i\) regularly spaced
evaluation checkpoints \(t_{i,1},\ldots,t_{i,M_i}\). Following the discrete
aggregation used in the Continual World result code
\citep{wolczyk2021continualworld}, the task-local acquisition score is
\begin{equation}
    \mathrm{AUC}_i
    =
    \frac{1}{M_i}
    \sum_{j=1}^{M_i}p_i(t_{i,j}).
\end{equation}
This is the mean height of the normalized learning curve while task \(i\) is
being trained. The reported curriculum score is the task macro-average
\begin{equation}
    \mathrm{AUC}
    =
    \frac{1}{N}\sum_{i=1}^{N}\mathrm{AUC}_i.
\end{equation}
Thus, AUC measures acquisition performance rather than retention after a task
block and lies in \([0,1]\).

\paragraph{Forward Transfer.}
Forward transfer compares each lifelong learning curve with a seed-matched
single-task reference trained from scratch
\citep{wolczyk2021continualworld}. In the FWT calculation, task-local areas
are estimated by the normalized trapezoidal rule. For a sequence
\(q_1,\ldots,q_M\), define
\begin{equation}
    \mathcal{A}(q_{1:M})
    =
    \frac{
        \tfrac{1}{2}q_1
        +\sum_{j=2}^{M-1}q_j
        +\tfrac{1}{2}q_M
    }{M-1}.
\end{equation}
Let \(A_i^{\mathrm{CL}}\) and \(A_i^{\mathrm{ST}}\) denote this area for
the lifelong and single-task learning curves, respectively. Then
\begin{equation}
    \mathrm{FWT}_i
    =
    \frac{
        A_i^{\mathrm{CL}}-A_i^{\mathrm{ST}}
    }{
        1-A_i^{\mathrm{ST}}
    },
    \qquad
    \mathrm{FWT}
    =
    \frac{1}{|\mathcal{I}_{\mathrm{FWT}}|}
    \sum_{i\in\mathcal{I}_{\mathrm{FWT}}}\mathrm{FWT}_i .
\end{equation}
The set \(\mathcal{I}_{\mathrm{FWT}}\) contains every task for which
\(1-A_i^{\mathrm{ST}}\geq10^{-3}\). This exclusion avoids unstable ratios
when the single-task reference is effectively saturated. No task is otherwise
removed from the reported FWT; in particular, CT28's four initial depth-2
tasks are retained. They are excluded only from selection analyses that
require an earlier hierarchy-relevant source.

Zero FWT matches single-task acquisition, positive values indicate faster
acquisition, and negative values indicate slower acquisition. The normalized
score is at most one but can be substantially negative when a strong
single-task reference appears in the denominator. AUC and FWT therefore need
not rank methods identically: AUC measures absolute acquisition, whereas FWT
first normalizes each task's improvement by the remaining headroom over its
single-task reference and then macro-averages the task scores. The reported
AUC and the trapezoidal area used by FWT also differ slightly in endpoint
weighting.

\paragraph{Forgetting.}
Continual World forgetting compares each task's performance at the end of its
training block with its performance at the final curriculum checkpoint
\citep{wolczyk2021continualworld}:
\begin{equation}
    \mathrm{FGT}_i
    =
    p_i(e_i)-p_i(T),
    \qquad
    \mathrm{FGT}
    =
    \frac{1}{N}\sum_{i=1}^{N}\mathrm{FGT}_i.
\end{equation}
Positive values indicate forgetting and negative values indicate subsequent
improvement. A value of zero indicates no net average change; complete
measured retention additionally requires every per-task difference to be
zero. The final task contributes zero because \(e_N=T\).

\section{Significance Testing}
\label{sec:significance_testing}

\paragraph{Statistical protocol.}
Deep reinforcement learning results vary across random seeds and environment stochasticity \citep{henderson2018deep, colas2018many}. Each comparison uses the seven shared seeds 86--92. We compute one metric value per seed, then test the mean paired difference \(\Delta=\mathrm{AMSC}-\mathrm{comparison}\) using an exact two-sided sign-flip test over all \(2^7=128\) sign assignments \citep{winkler2014permutation}. The tables also report 95\% percentile-bootstrap confidence intervals for the mean paired difference using \(10{,}000\) paired resamples with seed zero.

We report unadjusted \(p\)-values with 95\% confidence intervals. Green cells denote \(p<0.05\); red cells denote \(p\geq0.05\). Color indicates evidence for a difference,
not which method is better. Positive \(\Delta\) favors AMSC for AUC and FWT, whereas negative \(\Delta\) favors AMSC for FGT.

\paragraph{Baseline comparisons.}
Tables~\ref{tbl:pairwise_amsc_ct28}, \ref{tbl:pairwise_amsc_mg16}, and \ref{tbl:pairwise_amsc_cw10} use the principal configurations reported in the main text. CT28 uses \(N=128\); MG16 uses the seven completed \(N=1024,R=50,\rho=0.5\) runs from the sample-count sweep. With seven pairs, the smallest attainable two-sided raw \(p\)-value
is \(2/128=0.015625\).

\begin{table*}[h]
    \caption{CT28 baseline comparisons across seven matched seeds. Each metric reports the exact two-sided raw $p$ and a 95\% paired percentile-bootstrap CI for $\Delta=\mathrm{AMSC}-\mathrm{comparison}$. Green denotes $p<0.05$; red denotes $p\geq0.05$. Tests are not adjusted for multiple comparisons. Color indicates significance, not which method is better.}
    \label{tbl:pairwise_amsc_ct28}
    \centering
    \scriptsize
    \setlength{\tabcolsep}{2.5pt}
    \begin{tabular}{lcccccc}
        \toprule
        & \multicolumn{2}{c}{AUC} & \multicolumn{2}{c}{FWT} & \multicolumn{2}{c}{FGT} \\
        Method & $p$ & 95\% CI & $p$ & 95\% CI & $p$ & 95\% CI \\
        \midrule
        Fine-tuning & \cellcolor{green!10}0.016 & \cellcolor{green!10}$[0.56, 0.60]$ & \cellcolor{green!10}0.016 & \cellcolor{green!10}$[0.70, 0.74]$ & \cellcolor{green!10}0.016 & \cellcolor{green!10}$[-0.40, -0.36]$ \\
        From-scratch & \cellcolor{green!10}0.016 & \cellcolor{green!10}$[0.69, 0.75]$ & \cellcolor{green!10}0.016 & \cellcolor{green!10}$[0.91, 0.96]$ & \cellcolor{red!7}1.000 & \cellcolor{red!7}$[0.00, 0.00]$ \\
        SER-s & \cellcolor{red!7}0.422 & \cellcolor{red!7}$[-0.06, 0.02]$ & \cellcolor{red!7}0.578 & \cellcolor{red!7}$[-0.06, 0.02]$ & \cellcolor{green!10}0.016 & \cellcolor{green!10}$[-0.70, -0.36]$ \\
        SER-r & \cellcolor{red!7}0.188 & \cellcolor{red!7}$[0.00, 0.12]$ & \cellcolor{red!7}0.172 & \cellcolor{red!7}$[0.00, 0.14]$ & \cellcolor{red!7}0.125 & \cellcolor{red!7}$[-0.18, -0.02]$ \\
        SDW & \cellcolor{green!10}0.047 & \cellcolor{green!10}$[0.06, 0.28]$ & \cellcolor{green!10}0.047 & \cellcolor{green!10}$[0.07, 0.34]$ & \cellcolor{red!7}0.500 & \cellcolor{red!7}$[0.00, 0.02]$ \\
        OWL & \cellcolor{green!10}0.016 & \cellcolor{green!10}$[0.76, 0.82]$ & \cellcolor{green!10}0.016 & \cellcolor{green!10}$[1.09, 1.43]$ & \cellcolor{green!10}0.016 & \cellcolor{green!10}$[-0.21, -0.14]$ \\
        SANE & \cellcolor{green!10}0.016 & \cellcolor{green!10}$[0.44, 0.53]$ & \cellcolor{green!10}0.016 & \cellcolor{green!10}$[0.52, 0.61]$ & \cellcolor{red!7}1.000 & \cellcolor{red!7}$[0.00, 0.00]$ \\
        Mask-BLC & \cellcolor{green!10}0.016 & \cellcolor{green!10}$[0.63, 0.69]$ & \cellcolor{green!10}0.016 & \cellcolor{green!10}$[0.78, 0.89]$ & \cellcolor{red!7}1.000 & \cellcolor{red!7}$[0.00, 0.00]$ \\
        Mask-LC & \cellcolor{green!10}0.016 & \cellcolor{green!10}$[0.44, 0.49]$ & \cellcolor{green!10}0.016 & \cellcolor{green!10}$[0.53, 0.65]$ & \cellcolor{red!7}1.000 & \cellcolor{red!7}$[0.00, 0.00]$ \\
        Mask-LC (L2) & \cellcolor{green!10}0.016 & \cellcolor{green!10}$[0.37, 0.47]$ & \cellcolor{green!10}0.016 & \cellcolor{green!10}$[0.44, 0.58]$ & \cellcolor{red!7}1.000 & \cellcolor{red!7}$[0.00, 0.00]$ \\
        CKA-RL & \cellcolor{green!10}0.016 & \cellcolor{green!10}$[0.51, 0.64]$ & \cellcolor{green!10}0.016 & \cellcolor{green!10}$[0.60, 0.75]$ & \cellcolor{green!10}0.016 & \cellcolor{green!10}$[-0.41, -0.31]$ \\
        CompoNet & \cellcolor{green!10}0.016 & \cellcolor{green!10}$[0.15, 0.37]$ & \cellcolor{green!10}0.016 & \cellcolor{green!10}$[0.18, 0.43]$ & \cellcolor{red!7}1.000 & \cellcolor{red!7}$[0.00, 0.00]$ \\
        MOSAIC & \cellcolor{green!10}0.016 & \cellcolor{green!10}$[0.08, 0.17]$ & \cellcolor{green!10}0.016 & \cellcolor{green!10}$[0.11, 0.20]$ & \cellcolor{red!7}1.000 & \cellcolor{red!7}$[0.00, 0.00]$ \\
        MOSAIC \(\neg\mathbb{I}_{\mathrm{perf}}\) & \cellcolor{green!10}0.016 & \cellcolor{green!10}$[0.09, 0.17]$ & \cellcolor{green!10}0.016 & \cellcolor{green!10}$[0.11, 0.20]$ & \cellcolor{red!7}1.000 & \cellcolor{red!7}$[0.00, 0.00]$ \\
        \bottomrule
    \end{tabular}
\end{table*}

\begin{table*}[h]
    \caption{MG16 baseline comparisons across seven matched seeds. Each metric reports the exact two-sided raw $p$ and a 95\% paired percentile-bootstrap CI for $\Delta=\mathrm{AMSC}-\mathrm{comparison}$. Green denotes $p<0.05$; red denotes $p\geq0.05$. Tests are not adjusted for multiple comparisons. Color indicates significance, not which method is better.}
    \label{tbl:pairwise_amsc_mg16}
    \centering
    \scriptsize
    \setlength{\tabcolsep}{2.5pt}
    \begin{tabular}{lcccccc}
        \toprule
        & \multicolumn{2}{c}{AUC} & \multicolumn{2}{c}{FWT} & \multicolumn{2}{c}{FGT} \\
        Method & $p$ & 95\% CI & $p$ & 95\% CI & $p$ & 95\% CI \\
        \midrule
        Fine-tuning & \cellcolor{green!10}0.016 & \cellcolor{green!10}$[0.38, 0.50]$ & \cellcolor{green!10}0.016 & \cellcolor{green!10}$[0.33, 0.56]$ & \cellcolor{green!10}0.016 & \cellcolor{green!10}$[-0.41, -0.27]$ \\
        From-scratch & \cellcolor{green!10}0.016 & \cellcolor{green!10}$[0.45, 0.57]$ & \cellcolor{green!10}0.016 & \cellcolor{green!10}$[0.40, 0.66]$ & \cellcolor{red!7}1.000 & \cellcolor{red!7}$[0.00, 0.00]$ \\
        SER-s & \cellcolor{green!10}0.016 & \cellcolor{green!10}$[0.25, 0.36]$ & \cellcolor{green!10}0.016 & \cellcolor{green!10}$[0.50, 0.85]$ & \cellcolor{green!10}0.016 & \cellcolor{green!10}$[-0.60, -0.41]$ \\
        SER-r & \cellcolor{green!10}0.031 & \cellcolor{green!10}$[0.05, 0.16]$ & \cellcolor{green!10}0.016 & \cellcolor{green!10}$[0.23, 0.47]$ & \cellcolor{red!7}0.375 & \cellcolor{red!7}$[-0.18, 0.02]$ \\
        SDW & \cellcolor{green!10}0.016 & \cellcolor{green!10}$[0.14, 0.26]$ & \cellcolor{green!10}0.016 & \cellcolor{green!10}$[0.43, 0.62]$ & \cellcolor{red!7}0.094 & \cellcolor{red!7}$[-0.28, -0.03]$ \\
        OWL & \cellcolor{green!10}0.016 & \cellcolor{green!10}$[0.46, 0.59]$ & \cellcolor{green!10}0.016 & \cellcolor{green!10}$[0.67, 1.00]$ & \cellcolor{green!10}0.016 & \cellcolor{green!10}$[-0.24, -0.20]$ \\
        SANE & \cellcolor{green!10}0.016 & \cellcolor{green!10}$[0.31, 0.46]$ & \cellcolor{green!10}0.016 & \cellcolor{green!10}$[0.28, 0.43]$ & \cellcolor{red!7}1.000 & \cellcolor{red!7}$[0.00, 0.00]$ \\
        Mask-BLC & \cellcolor{green!10}0.016 & \cellcolor{green!10}$[0.09, 0.27]$ & \cellcolor{green!10}0.016 & \cellcolor{green!10}$[0.17, 0.45]$ & \cellcolor{red!7}1.000 & \cellcolor{red!7}$[0.00, 0.00]$ \\
        Mask-LC & \cellcolor{green!10}0.016 & \cellcolor{green!10}$[0.20, 0.45]$ & \cellcolor{green!10}0.016 & \cellcolor{green!10}$[0.44, 1.10]$ & \cellcolor{red!7}1.000 & \cellcolor{red!7}$[0.00, 0.00]$ \\
        Mask-LC (L2) & \cellcolor{green!10}0.016 & \cellcolor{green!10}$[0.21, 0.35]$ & \cellcolor{green!10}0.016 & \cellcolor{green!10}$[0.16, 0.35]$ & \cellcolor{red!7}1.000 & \cellcolor{red!7}$[0.00, 0.00]$ \\
        CKA-RL & \cellcolor{green!10}0.016 & \cellcolor{green!10}$[0.34, 0.47]$ & \cellcolor{green!10}0.016 & \cellcolor{green!10}$[0.64, 0.87]$ & \cellcolor{green!10}0.016 & \cellcolor{green!10}$[-0.54, -0.28]$ \\
        CompoNet & \cellcolor{green!10}0.016 & \cellcolor{green!10}$[0.34, 0.51]$ & \cellcolor{green!10}0.016 & \cellcolor{green!10}$[1.18, 1.63]$ & \cellcolor{red!7}1.000 & \cellcolor{red!7}$[0.00, 0.00]$ \\
        MOSAIC & \cellcolor{green!10}0.016 & \cellcolor{green!10}$[0.35, 0.53]$ & \cellcolor{green!10}0.016 & \cellcolor{green!10}$[0.72, 1.38]$ & \cellcolor{red!7}1.000 & \cellcolor{red!7}$[0.00, 0.00]$ \\
        MOSAIC \(\neg\mathbb{I}_{\mathrm{perf}}\) & \cellcolor{green!10}0.016 & \cellcolor{green!10}$[0.34, 0.56]$ & \cellcolor{green!10}0.016 & \cellcolor{green!10}$[0.74, 1.54]$ & \cellcolor{red!7}1.000 & \cellcolor{red!7}$[0.00, 0.00]$ \\
        \bottomrule
    \end{tabular}
\end{table*}

\paragraph{CW10 comparisons.}
CW10 uses seed-matched online success evaluations. AMSC's FWT is lower than Mask-BLC (\(p=0.031\)) and from-scratch learning (\(p=0.016\)) at the nominal 0.05 threshold. The AUC difference from Mask-BLC is not significant (\(p=0.328\)). FWT excludes task--seed reference headroom below \(10^{-3}\); near-saturated reference curves can still amplify negative values.

\begin{table*}[h]
    \caption{CW10 baseline comparisons across seven matched seeds. ProgNet uses its native two-layer scratch FWT reference; other methods use the four-layer reference. Each metric reports the exact two-sided raw $p$ and a 95\% paired percentile-bootstrap CI for $\Delta=\mathrm{AMSC}-\mathrm{comparison}$. Green denotes $p<0.05$; red denotes $p\geq0.05$. Tests are not adjusted for multiple comparisons. Color indicates significance, not which method is better.}
    \label{tbl:pairwise_amsc_cw10}
    \centering
    \scriptsize
    \setlength{\tabcolsep}{2.5pt}
    \begin{tabular}{lcccccc}
        \toprule
        & \multicolumn{2}{c}{AUC} & \multicolumn{2}{c}{FWT} & \multicolumn{2}{c}{FGT} \\
        Method & $p$ & 95\% CI & $p$ & 95\% CI & $p$ & 95\% CI \\
        \midrule
        Fine-tuning & \cellcolor{red!7}0.234 & \cellcolor{red!7}$[-0.04, 0.35]$ & \cellcolor{red!7}0.703 & \cellcolor{red!7}$[-1.15, 18.58]$ & \cellcolor{red!7}0.062 & \cellcolor{red!7}$[-0.58, -0.18]$ \\
        From-scratch & \cellcolor{red!7}0.062 & \cellcolor{red!7}$[0.02, 0.12]$ & \cellcolor{green!10}0.016 & \cellcolor{green!10}$[-2.19, -0.43]$ & \cellcolor{red!7}0.500 & \cellcolor{red!7}$[-0.01, 0.00]$ \\
        ProgNet & \cellcolor{green!10}0.016 & \cellcolor{green!10}$[0.08, 0.21]$ & \cellcolor{red!7}0.141 & \cellcolor{red!7}$[-1.90, -0.06]$ & \cellcolor{red!7}0.500 & \cellcolor{red!7}$[-0.01, 0.00]$ \\
        Mask-LC & \cellcolor{red!7}0.781 & \cellcolor{red!7}$[-0.08, 0.10]$ & \cellcolor{red!7}0.062 & \cellcolor{red!7}$[-2.01, -0.22]$ & \cellcolor{red!7}1.000 & \cellcolor{red!7}$[-0.01, 0.01]$ \\
        Mask-BLC & \cellcolor{red!7}0.328 & \cellcolor{red!7}$[-0.12, 0.03]$ & \cellcolor{green!10}0.031 & \cellcolor{green!10}$[-2.21, -0.58]$ & \cellcolor{red!7}0.500 & \cellcolor{red!7}$[-0.01, 0.00]$ \\
        CKA-RL & \cellcolor{red!7}0.141 & \cellcolor{red!7}$[0.00, 0.20]$ & \cellcolor{red!7}0.453 & \cellcolor{red!7}$[-0.59, 2.77]$ & \cellcolor{red!7}0.500 & \cellcolor{red!7}$[-0.01, 0.00]$ \\
        MOSAIC & \cellcolor{green!10}0.016 & \cellcolor{green!10}$[0.16, 0.29]$ & \cellcolor{green!10}0.016 & \cellcolor{green!10}$[0.49, 9.03]$ & \cellcolor{red!7}0.500 & \cellcolor{red!7}$[-0.01, 0.00]$ \\
        \bottomrule
    \end{tabular}
\end{table*}

\paragraph{Composition and retrieval controls.}
Table~\ref{tbl:pairwise_amsc_ablation} compares each control with
configuration-matched AMSC runs across seven shared seeds.
CT28 uses \(N=128\), while all MG16 comparisons use
\(N=1024\), \(R=50\), and \(\rho=0.5\).
Each control changes only the specified retrieval or composition component.
FGT is omitted because all controls and their matched AMSC references
have zero measured forgetting.

\begin{table*}[h]
    \caption{CT28 and MG16 paired ablation comparisons across seven matched seeds. Each metric reports the exact two-sided unadjusted $p$-value and a 95\% paired bootstrap CI for $\Delta=\mathrm{AMSC}-\mathrm{control}$. Green denotes $p<0.05$; red denotes $p\geq0.05$. Positive differences favor AMSC. CT28 uses $N=128$. MG16 uses $N=1024$, with $R=50$ and $\rho=0.5$.}
    \label{tbl:pairwise_amsc_ablation}
    \centering
    \scriptsize
    \setlength{\tabcolsep}{2.5pt}
    \begin{tabular}{lcc cc cc cc}
        \toprule
        & \multicolumn{4}{c}{CT28} & \multicolumn{4}{c}{MG16} \\
        \cmidrule(lr){2-5}\cmidrule(lr){6-9}
        & \multicolumn{2}{c}{AUC} & \multicolumn{2}{c}{FWT} & \multicolumn{2}{c}{AUC} & \multicolumn{2}{c}{FWT} \\
        \cmidrule(lr){2-3}\cmidrule(lr){4-5}\cmidrule(lr){6-7}\cmidrule(lr){8-9}
        Configuration & $p$ & 95\% CI & $p$ & 95\% CI & $p$ & 95\% CI & $p$ & 95\% CI \\
        \midrule
        Learned \(\beta\) & \cellcolor{red!7}0.484 & \cellcolor{red!7}$[-0.02, 0.06]$ & \cellcolor{red!7}0.453 & \cellcolor{red!7}$[-0.02, 0.08]$ & \cellcolor{green!10}0.031 & \cellcolor{green!10}$[0.04, 0.23]$ & \cellcolor{green!10}0.047 & \cellcolor{green!10}$[0.06, 0.38]$ \\
        Uniform & \cellcolor{red!7}0.172 & \cellcolor{red!7}$[-0.01, 0.09]$ & \cellcolor{red!7}0.172 & \cellcolor{red!7}$[-0.01, 0.11]$ & \cellcolor{green!10}0.047 & \cellcolor{green!10}$[0.04, 0.20]$ & \cellcolor{green!10}0.016 & \cellcolor{green!10}$[0.05, 0.21]$ \\
        No \(\ell_2\)-STE & \cellcolor{green!10}0.047 & \cellcolor{green!10}$[0.01, 0.08]$ & \cellcolor{green!10}0.016 & \cellcolor{green!10}$[0.04, 0.11]$ & \cellcolor{red!7}0.734 & \cellcolor{red!7}$[-0.10, 0.16]$ & \cellcolor{red!7}0.172 & \cellcolor{red!7}$[-0.05, 0.36]$ \\
        Frozen retrieval & \cellcolor{green!10}0.047 & \cellcolor{green!10}$[0.03, 0.12]$ & \cellcolor{red!7}0.062 & \cellcolor{red!7}$[0.03, 0.14]$ & \cellcolor{green!10}0.031 & \cellcolor{green!10}$[0.02, 0.22]$ & \cellcolor{green!10}0.016 & \cellcolor{green!10}$[0.05, 0.22]$ \\
        Top-one & \cellcolor{red!7}0.078 & \cellcolor{red!7}$[0.02, 0.11]$ & \cellcolor{red!7}0.078 & \cellcolor{red!7}$[0.02, 0.12]$ & \cellcolor{red!7}0.844 & \cellcolor{red!7}$[-0.12, 0.09]$ & \cellcolor{red!7}0.844 & \cellcolor{red!7}$[-0.16, 0.10]$ \\
        Matched random & \cellcolor{green!10}0.016 & \cellcolor{green!10}$[0.56, 0.62]$ & \cellcolor{green!10}0.016 & \cellcolor{green!10}$[0.74, 0.85]$ & \cellcolor{green!10}0.016 & \cellcolor{green!10}$[0.17, 0.34]$ & \cellcolor{green!10}0.016 & \cellcolor{green!10}$[0.18, 0.39]$ \\
        Transfer-informed & \cellcolor{green!10}0.016 & \cellcolor{green!10}$[0.10, 0.15]$ & \cellcolor{green!10}0.016 & \cellcolor{green!10}$[0.15, 0.33]$ & \cellcolor{red!7}0.531 & \cellcolor{red!7}$[-0.14, 0.05]$ & \cellcolor{green!10}0.031 & \cellcolor{green!10}$[-0.27, -0.08]$ \\
        \bottomrule
    \end{tabular}
\end{table*}


\subsection{Pairwise Transfer Matrices}
\label{sec:pairwise_transfer_matrices}

\begin{figure*}[p]
    \centering
    \includegraphics[width=0.98\textwidth]{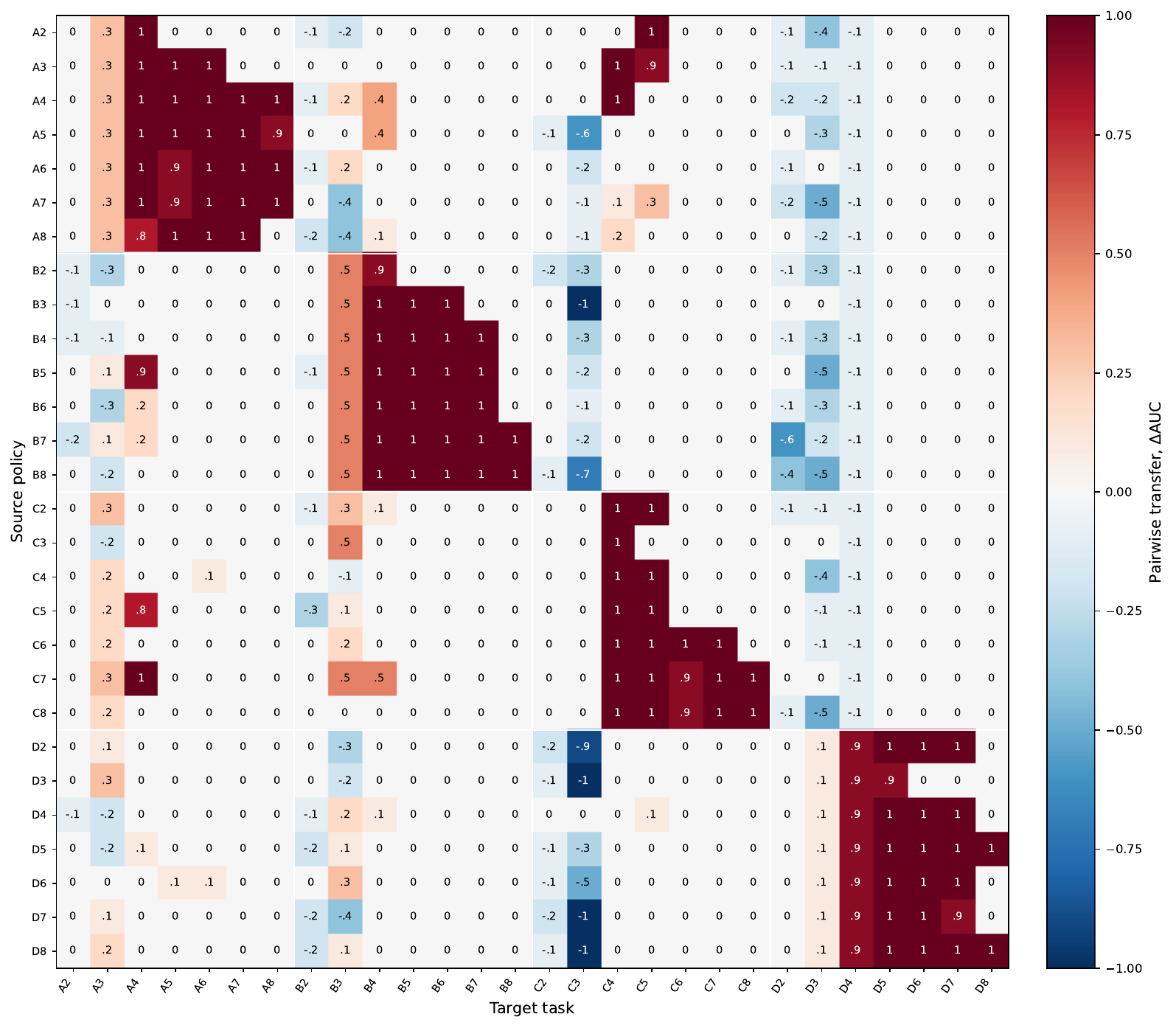}
    \caption{Raw pairwise transfer on CT28. Target training is initialized from
    the source policy and then fine-tuned. Cell annotations report
    \(\Delta\mathrm{AUC}\) relative to learning the target without a source;
    red denotes positive transfer and blue denotes negative transfer. Source
    policies are taken from the family-isolated seed-92 runs. Family labels
    A--D and depths 2--8 define the row and column order.}
    \label{fig:ct28_pairwise_transfer_matrix}
\end{figure*}

\begin{figure*}[p]
    \centering
    \includegraphics[width=0.98\textwidth]{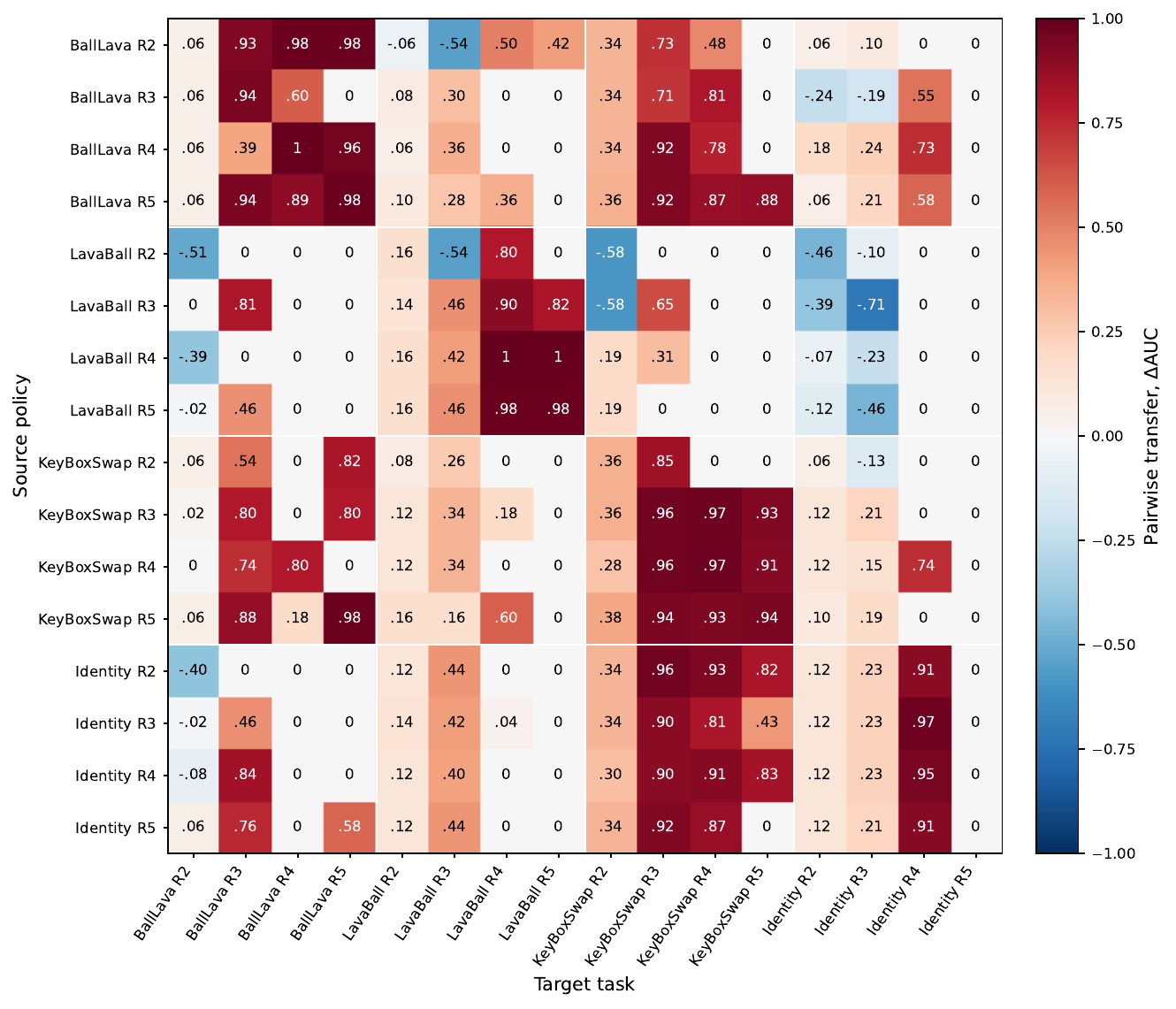}
    \caption{Raw pairwise transfer on MG16. Each source mask is composed with a
    newly initialized target mask using an initially equal mixture, and both
    the target mask and mixture coefficients are optimized on the target task.
    Cell annotations report \(\Delta\mathrm{AUC}\) relative to the matched
    no-source run; red denotes positive transfer and blue denotes negative
    transfer. Source masks are taken from the family-isolated seed-86 runs.
    Tasks are grouped by observation-remapping family and ordered by radius
    2--5.}
    \label{fig:mg16_pairwise_transfer_matrix}
\end{figure*}

Figures~\ref{fig:ct28_pairwise_transfer_matrix} and
\ref{fig:mg16_pairwise_transfer_matrix} report the unaggregated pairwise
transfer measurements used in the retrieval analysis. Each cell is
\(\Delta\mathrm{AUC}\), the target-task AUC obtained with the row source minus
the AUC of the corresponding no-source run. Positive values indicate improved
learning over the target budget and negative values indicate interference. Rows are source
policies and columns are target tasks; tasks are grouped by family and ordered
by increasing depth or radius. The complete matrices include self-pairs and
sources that occur after the target in the lifelong curriculum. The curriculum-restricted
analysis in Section~\ref{sec:selection_analysis} retains only sources available
when the target is encountered.

\subsection{Per-Task Policy-Mask Utilization}
\label{sec:mask_utilization}

\begin{figure*}[p]
    \centering
    \includegraphics[width=0.98\textwidth]{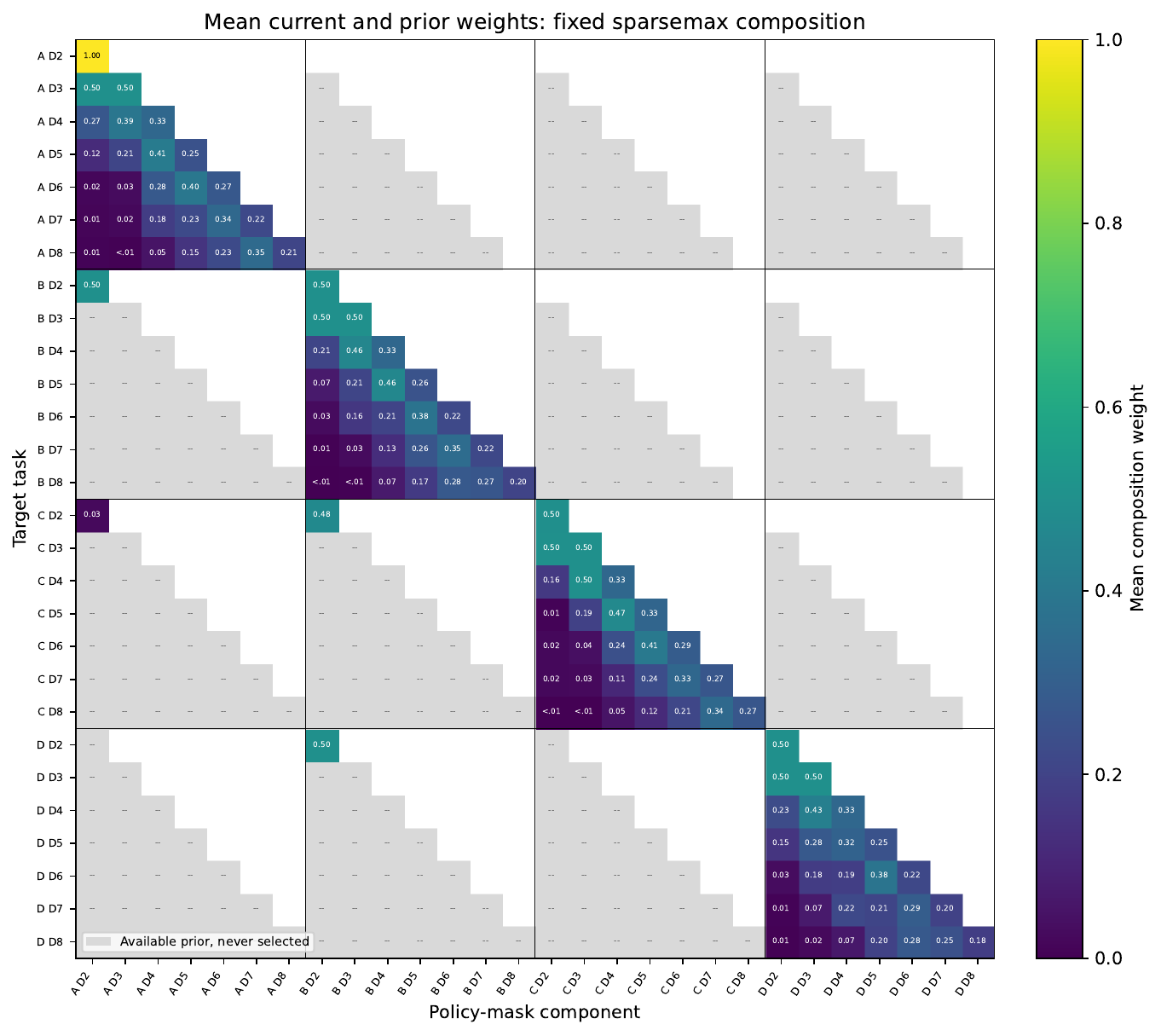}
    \caption{Per-task policy-mask utilization on CT28. Cell annotations report
    the mean composition coefficient assigned to each current or prior mask.
    The block structure shows that AMSC places most prior mass on masks from
    the same task family, while retaining limited cross-family reuse where the
    online representation retrieves it.}
    \label{fig:ct28_mask_utilization}
\end{figure*}

\begin{figure*}[p]
    \centering
    \includegraphics[width=0.98\textwidth]{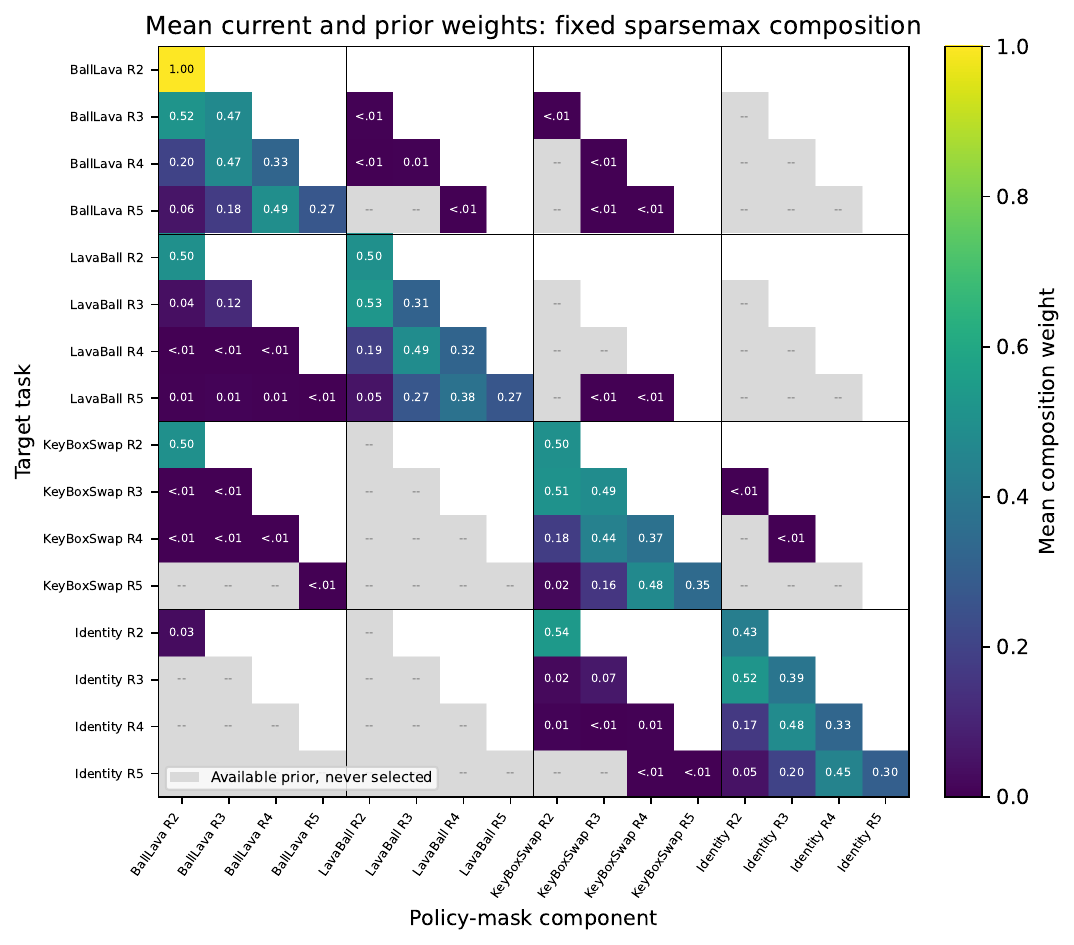}
    \caption{Per-task policy-mask utilization on MG16. Cell annotations report
    the mean composition coefficient assigned to each current or prior mask.
    In contrast to CT28, AMSC assigns non-negligible mass to several
    cross-family components, consistent with the less family-separable
    transfer structure of MG16.}
    \label{fig:mg16_mask_utilization}
\end{figure*}

Figures~\ref{fig:ct28_mask_utilization} and
\ref{fig:mg16_mask_utilization} report how AMSC distributes composition mass
across the current and retrieved policy masks. Rows are target tasks and
columns are policy-mask components. Diagonal cells therefore correspond to
the current mask, while colored cells below the diagonal correspond to prior
masks available at that point in the curriculum. Values are the mean fixed
composition coefficients across retrieval events and seven seeds. Grey cells
identify available priors that are never selected; blank cells correspond to
future masks that are not yet available. These coefficients quantify mask
reuse, rather than the fraction of backbone parameters activated by a mask.

\subsection{Initial Retrieval and Temporal Refinement}
\label{sec:temporal_retrieval}

We compare AMSC's first eligible retrieval and its final retrieval with the
pairwise-transfer measurements above. The first event occurs after the
current-task experience buffer contains enough samples to construct an
embedding. We exclude task runs with fewer than two measured candidate sources because
they contain no meaningful source-selection decision.

Table~\ref{tbl:temporal_retrieval_alignment} reports alignment with measured
transfer. Positive-transfer precision is the fraction of selected sources with
\(\Delta\mathrm{AUC}>0\). Positive-mass recall is given by
\begin{equation}
    \frac{
        \sum_{i\in\mathcal{P}(k)}
        \max(\Delta\mathrm{AUC}_{i\rightarrow k},0)
    }{
        \sum_{i<k}
        \max(\Delta\mathrm{AUC}_{i\rightarrow k},0)
    },
\end{equation}
which weights recovered sources by transfer magnitude rather than counting all
positive pairs equally. Best-source inclusion indicates whether the support
contains a source attaining the largest available \(\Delta\mathrm{AUC}\).
Parenthesized values are analytic expectations under uniform sampling without
replacement of a support with the same size.

\begin{table*}[t]
    \caption{Alignment of initial and final AMSC retrieval with aggregate
    pairwise transfer. Weighted \(\Delta\mathrm{AUC}\) normalizes weights over
    retrieved sources and therefore excludes current-mask mass. Parentheses
    give support-size-matched random expectations. Statistics aggregate seven
    lifelong runs; the transfer value assigned to each pair comes from one
    source-policy training seed.}
    \label{tbl:temporal_retrieval_alignment}
    \begin{center}
    \scriptsize
    \begin{tabular}{llrrrrrr}
        \toprule
        Benchmark & Phase & Support
        & Mean \(\Delta\)AUC
        & Weighted \(\Delta\)AUC
        & Positive precision
        & Positive-mass recall
        & Best source \\
        \midrule
        CT28 & Initial & 3.11 & 0.52 (0.10) & 0.49
             & 0.61 (0.19) & 0.87 (0.23) & 0.98 (0.47) \\
             & Final   & 2.57 & 0.57 (0.10) & 0.62
             & 0.67 (0.19) & 0.86 (0.20) & 0.96 (0.43) \\
        MG16 & Initial & 2.42 & 0.47 (0.20) & 0.51
             & 0.74 (0.46) & 0.63 (0.33) & 0.84 (0.41) \\
             & Final   & 1.94 & 0.53 (0.20) & 0.55
             & 0.80 (0.46) & 0.58 (0.27) & 0.84 (0.36) \\
        \bottomrule
    \end{tabular}
    \end{center}
\end{table*}

The initial support is already strongly transfer-aligned. Its mean
\(\Delta\mathrm{AUC}\) exceeds the random expectation by \(0.42\) on
CT28 and \(0.27\) on MG16; seed-bootstrap 95\% confidence intervals are
\([0.40,0.43]\) and \([0.25,0.29]\). Initial retrieval also contains the best
available source in \(98\%\) and \(84\%\) of task runs. Refinement trades some
positive-transfer mass recall for greater precision, consistent with pruning
an initially over-inclusive support.

\begin{table*}[t]
    \caption{Temporal retrieval diagnostics. \(J\) denotes support Jaccard
    similarity. Improved and worsened are fractions among task runs whose
    first and final supports differ, measured by source-normalized,
    similarity-weighted \(\Delta\mathrm{AUC}\).}
    \label{tbl:temporal_retrieval_refinement}
    \begin{center}
    \scriptsize
    \begin{tabular}{lrrrrrrr}
        \toprule
        Benchmark
        & Consecutive \(J\)
        & First--final \(J\)
        & Changed
        & Support
        & Weighted \(\Delta\)AUC
        & Improved
        & Worsened \\
        \midrule
        CT28 & 0.97 & 0.82 & 0.43 & \(3.11\rightarrow2.57\)
             & \(0.49\rightarrow0.62\) & 0.70 & 0.23 \\
        MG16 & 0.97 & 0.82 & 0.39 & \(2.42\rightarrow1.94\)
             & \(0.51\rightarrow0.55\) & 0.74 & 0.16 \\
        \bottomrule
    \end{tabular}
    \end{center}
\end{table*}

Although only \(10\%\) of consecutive CT28 retrieval events and \(7\%\) of
MG16 events change support, these changes accumulate over training. Across
seeds, CT28 support changes add 23 and remove 122 source occurrences; their
mean \(\Delta\mathrm{AUC}\) values are \(0.39\) and \(0.16\), respectively.
MG16 changes add 8 and remove 55 occurrences, with corresponding means of
\(0.28\) and \(0.22\). The source-normalized weighted improvement from first
to final retrieval is \(0.13\) on CT28 (95\% CI \([0.11,0.15]\)) and \(0.03\)
on MG16 (\([0.01,0.05]\)).

These diagnostics associate online refinement with greater aggregate
pairwise transfer; they do not provide time-local transfer labels. A
No Periodicity control tests repeated retrieval by retaining the first valid
support and weights throughout each task. On CT28, AUC/FWT falls from
\(0.94/0.94\) to \(0.87/0.85\), with paired \(p=0.047\) and \(p=0.062\).

On MG16, retaining the first retrieval lowers AUC/FWT from \(0.73/0.54\) to \(0.63/0.42\), with paired (\(p=0.031\)) and (\(p=0.016\)), respectively, using matched (\(N=1024\)) configurations. These results support updating retrieval as target-task experience accumulates. The pairwise transfer measurements summarize complete target learning curves from one source-policy seed, so their alignment with retrieval should be interpreted as a transfer proxy rather than direct evidence of instantaneous benefit.


\end{document}